\documentclass{article}

\usepackage[preprint]{neurips_2025}

\usepackage[utf8]{inputenc} 
\usepackage[T1]{fontenc}    
\usepackage{hyperref}       
\usepackage{url}            
\usepackage{booktabs}       
\usepackage{amsfonts}       
\usepackage{minitoc}        
\usepackage{titletoc}
\usepackage[ruled,vlined]{algorithm2e} 
\usepackage{nicefrac}       
\usepackage{microtype}      
\usepackage{xcolor}         
\usepackage{graphicx} 
\usepackage{enumitem}
\usepackage{minitoc}

\usepackage{amsmath}
\usepackage{amssymb}
\usepackage{mathtools}
\usepackage{amsthm}
\usepackage{multirow}
\usepackage{multicol}
\theoremstyle{plain}
\newtheorem{theorem}{Theorem}[section]

\newtheorem{definition}[theorem]{Definition}

\definecolor{royalblue}{rgb}{0.25, 0.41, 0.88}
\hypersetup{
    colorlinks=true,
    citecolor=royalblue,
    linkcolor=royalblue,  
    filecolor=magenta,      
    urlcolor=royalblue
}

\usepackage{enumitem}
\usepackage[most]{tcolorbox}
\tcbset{
    enhanced,        
    breakable,          
    colback=yellow!20 
    ,colframe=yellow!50!black    
    ,boxrule=1pt     
    ,arc=5pt          
    ,auto outer arc   
    ,boxsep=5pt       
}
\title{A Game-Theoretic Negotiation Framework for Cross-Cultural Consensus in LLMs}

\author{%
    Guoxi Zhang\thanks{Equal contribution, $^{\dag}$corresponding author. Project website: \url{https://pku-culture-negotiation.github.io}.}\;\;
    Jiawei Chen$^{*}$\;\;
    Tianzhuo Yang$^{*}$\;\;
    Jiaming Ji\;\;
    Yaodong Yang$^{\dag}$\;\;
    \textbf{Juntao Dai}$^{\dag}$\\
    \vspace{-0.5em} \\
    \textnormal{Institute for Artificial Intelligence}\\
    \textnormal{Peking University}
}

\begin{document}

\maketitle

\begin{abstract}
The increasing prevalence of large language models (LLMs) is influencing global value systems. However, these models frequently exhibit a pronounced WEIRD (Western, Educated, Industrialized, Rich, Democratic) cultural bias due to lack of attention to minority values. This monocultural perspective may reinforce dominant values and marginalize diverse cultural viewpoints,  posing challenges for the development of equitable and inclusive AI systems. In this work, we introduce a systematic framework designed to boost fair and robust cross-cultural consensus among LLMs. We model consensus as a Nash Equilibrium and employ a game-theoretic negotiation method based on Policy-Space Response Oracles (PSRO) to simulate an organized cross-cultural negotiation process. To evaluate this approach, we construct regional cultural agents using data transformed from the World Values Survey (WVS). Beyond the conventional model-level evaluation method, We further propose two quantitative metrics, Perplexity-based Acceptence and Values Self-Consistency, to assess consensus outcomes. Experimental results indicate that our approach generates consensus of higher quality while ensuring more balanced compromise compared to baselines. Overall, it mitigates WEIRD bias by guiding agents toward convergence through fair and gradual negotiation steps. 
\end{abstract}

\section{Introduction}
\label{sec:intro}
The widespread adoption of large language models (LLMs) is reshaping global social values. However, these models often exhibit a pronounced WEIRD bias, favoring Western, Educated, Industrialized, Rich and Democratic perspectives~\cite{agarwalEthicalReasoningMoral2024, benklerAssessingLLMsMoral2023, liuCulturalBiasLarge2024, quPerformanceBiasesLarge2024}. As LLMs become increasingly embedded in policy-making and public governance~\cite{aoki2024largelanguagemodelspolitics, jiangEditorialLargeLanguage2025}, this monocultural orientation risks the domination of prevailing social values and the \textit{lock-in} of controversial moral beliefs across broader contexts~\cite{liuCulturalBiasLarge2024, qiuProgressGymAlignmentMillennium2024}.

Enabling equitable dialogue and effective negotiation among diverse cultures within AI systems has therefore become a growing concern in global AI governance~\cite{oheigeartaighOvercomingBarriersCrosscultural2020, gurkan2024exploringpublicopinionresponsible}. The establishment of cultural consensus forms a basis for resolving cross-cultural conflicts and supporting international cooperation. Given the complexity of multicultural scenarios, there is an urgent need to develop automated \textit{cultural consensus solvers} to facilitate consensus-building among diverse cultural perspectives.

Achieving cross-cultural consensus, however, presents several challenges. First, the lack of fined culture-alignment methods often results in models defaulting to superficial \textit{value labeling} or one-sided cultural representations~\cite{benklerAssessingLLMsMoral2023, parkMoralityNonbinaryBuilding2024, zhouRethinkingMachineEthics2024}. Second, existing approaches like debate protocols typically rely on random interactions and majority voting, which do not ensure fairness in the consensus process~\cite{zhang2025multiagentdebateanswerquestion}. Our experiments show that conventional debate mechanisms often assimilate less-represented cultures into dominant WEIRD value systems, producing implicit value domination, distorting consensus outcomes and worsening cross-cultural negotiation imbalances.  Third, there is a lack of quantitative standards for evaluating the quality of consensus achieved.

To address these challenges, we present a systematic framework for reaching cross-cultural consensus. We first introduce a game-theoretic formulation of consensus as a Nash Equilibrium and design a PSRO-based consensus-solving method to enable fair negotiation among diverse cultural agents. Building on this, we propose a culture-anchoring approach for precise modeling of individual cultural groups. Finally, we develop new quantitative metrics to comprehensively evaluate both the negotiation processes and the outcomes between different cultural agents.

Our main contribution is the game-theoretic framework consisting of three parts listed as follows:
\begin{itemize}[leftmargin=*]
    \item \textbf{Cross-Cultural Negotiation:} We define cultural consensus from a game-theoretic perspective and propose a PSRO-based negotiation method to facilitate fair and robust agreement. This approach provides theoretical guarantees of fairness and procedural justice in consensus-building, and generates high-quality, globally-applicable AI alignment data.
    \item \textbf{Regional Cultural Agents:} To validate our method, we systematically construct and evaluate eight culturally-aligned agents based on WVS and Hofstede's Culture Dimensions Theory, qualifiying as representitive negotiation participants for targeted cultures.
    \item \textbf{Consensus Evaluation Toolkit:} To address the lack of consensus evaluation standards, we introduce two quantitative metrics for consensus assessment, Perplexity-based Acceptence and Values Self-Consistency, revealing limitations of traditional baselines and systematically validating the effectiveness of our approach in real-world multicultural scenarios.
\end{itemize}

\section{Related Work}

\begin{figure}
    \centering
    \includegraphics[width=1\linewidth]{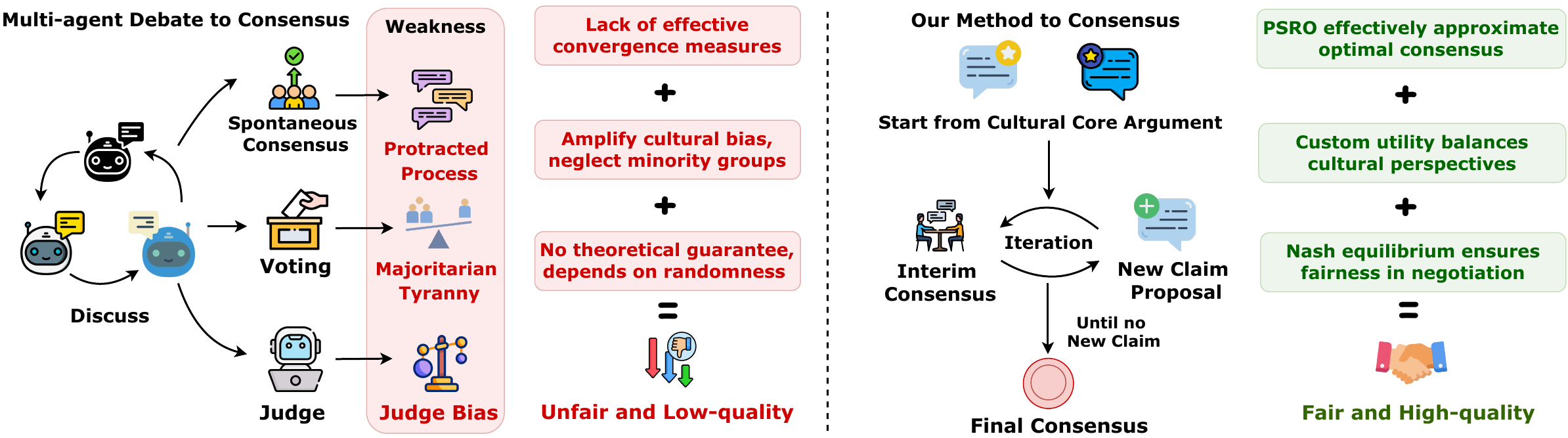}
    \caption{\textbf{Comparison of traditional debate-based consensus methods and our method.} Traditional methods (judge, voting, spontaneous consensus) suffer from bias, unfairness, and lack of convergence guarantees. Our approach starts from core cultural arguments, using PSRO with custom utility functions to reach a fair, Nash Equilibrium-based cultural consensus.}
    \label{fig:debate_methods}
\end{figure}

\noindent{\textbf{Value Theories and Alignment\quad}} Several established frameworks provide the foundation for cross-cultural value assessment. The World Values Survey (WVS)~\cite{WVS_Wave7} examines how human values relates to social and political development across over 120 societies. Building on this, the Inglehart-Welzel Cultural Map offers a two-dimensional model of cultural variation ~\cite{OughtComparisonUnsupervised2024,kabir2025breakcheckboxchallengingclosedstyle}. Hofstede's Cultural Dimensions Theory (VSM13)~\cite{kharchenko2024llmsrepresentvaluescultures,masoud2024culturalalignmentlargelanguage,masoud2025culturalalignmentlargelanguage} provides a standardized six-dimensional framework for measuring cultural traits~\cite{wang2024cdevalbenchmarkmeasuringcultural}. Schwartz's Theory of Basic Values~\cite{schwartzUniversalsContentStructure1992} organizes ten core values along two bipolar dimensions, and has been adopted to evaluate the values of LLMs~\cite{yao2023valuefulcramappinglarge}. These theories are further detailed in Appendix~\ref{appendix:tool}. Some works focus on region-specific value alignment~\cite{huangAceGPTLocalizingLarge2024, linTaiwanLLMBridging2023}. CultureBench emphasizes cultural commonsense evaluation ~\cite{chiu2024culturalbenchrobustdiversechallenging}, providing complementary approaches to measuring how well AI systems represent diverse cultural perspectives.

\noindent{\textbf{Multi-Agent Debate (MAD) and Game Theory\quad}} MAD has been shown to improve LLMs reasoning by integrating diverse agent feedbacks
~\cite{khanDebatingMorePersuasive2024}. In the context of cultural conflict, MAD allows different cultural perspectives to interact and potentially reach consensus through deliberation. Typical debate protocols include emergent consensus via iterative dialogue~\cite{duImprovingFactualityReasoning2023}, judge-based evaluation~\cite{liangEncouragingDivergentThinking2024} and majority voting~\cite{yangLLMVotingHuman2024}, as well as more recent variants like role-play~\cite{wangUnleashingEmergentCognitive2024, chanChatEvalBetterLLMbased2023, pangSelfAlignmentLargeLanguage2024} and subgroup discussion~\cite{wangRethinkingBoundsLLM2024, maiSequenceGraphNetwork2025}. However, these methods face limitations: voting and judge-based protocols can amplify model bias or introduce value contamination~\cite{zhang2025multiagentdebateanswerquestion, zhao2024electoralapproachdiversifyllmbased}, while emergent consensus may result in negotiation deadlocks~\cite{zhang2025multiagentdebateanswerquestion}. To address these issues, game theory provides a more quantifiable foundation~\cite{ hua2024gametheoreticllmagentworkflow, dudikGametheoreticApproachModeling2013}. Recent work, such as the \textit{consensus game} framework, models LLMs interactions as equilibrium search problems to promote robust consensus~\cite{jacobConsensusGameLanguage2023}. In practice, due to the vastness of the argument strategy space, methods like Policy-Space Response Oracles (PSRO) are used to iteratively expand the candidate strategy set and search for equilibria~\cite{lanctotUnifiedGameTheoreticApproach2017}, providing a method for more rigorous consensus achievement.

\section{Cross-Cultural Negotiation}
\label{sec:method}
Our definition of cultural negotiation is informed by theories of deliberative democracy~\cite{Cohen2009-COHPPD, Gutmann2004-GUTWDD-2}, which conceptualize the process as structured, iterative and oriented toward legitimate consensus through rational discourse and mutual adjustment. Building on this foundation, we formalize the cultural negotiation problem as a two-player game, explicitly defining utility and consensus to achieve the balance between core values and compromise. We then design a negotiation process based on PSRO~\cite{lanctotUnifiedGameTheoreticApproach2017}. This approach enables agents to systematically search for fair and robust consensus by repeatedly proposing and adjusting culturally grounded strategies.

\subsection{Formalization}

Formally, we model the cultural negotiation process as a two-player extensive-form game, represented by the quintuple:
$\Gamma \doteq \langle \mathcal{I}, \mathcal{G}, \mathcal{W}, \mathcal{U}, \mathcal{H} \rangle$
, where:
\begin{itemize}[leftmargin=*]
\item \textbf{Cultural Entities:} $\mathcal{I} \doteq \{A, B\}$, the set of two distinct cultural entities involved in the negotiation, where $A$ and $B$ represent different cultures with their own values and perspectives.
\item \textbf{Guideline Sets:} $\mathcal{G} \doteq \{G_i | i \in \mathcal{I}\}$, each guideline $g \in G_i$ is structured as a triple $g = \langle \text{content}, \text{reason}, \text{description} \rangle$, capturing the natural language specification of core cultural imperatives on specific topics.
\item \textbf{Guideline Weights:} $\mathcal{W} \doteq \{W_i | i \in \mathcal{I}\}$, for each culture $i \in \mathcal{I}$, $W_i \in \Delta(G_i)$ denotes a probability distribution over its guidelines, with $\sum_g w_i(g) = 1$. $W_i$ thus characterizes the expressive emphasis of culture $i$ in the current negotiation round.
\item \textbf{Utility Functions:} $\mathcal{U} \doteq \{U_i | i \in \mathcal{I}\}$, quantify the utility each culture derives from different guideline combinations.
\item \textbf{Negotiation History:} $\mathcal{H}$, the sequence of utterances and proposals exchanged in negotiation.\end{itemize}

\subsection{Utility}
Drawing on the theory of \textit{overlapping consensus}~\cite{Rawls1993-RAWPL}, we define utility on two primary components: \textbf{Consistency}, which measures the extent to which a cultural entity maintains its core principles and \textbf{Acceptance}, which measures the degree to which its proposals are acceptable to the other party. To address issues observed in debate settings, such as repetitive argumentation and diminished quality, we introduce a \textbf{Novelty} component that penalizes redundancy and encourages innovation. The necessity of incorporating Novelty is demonstrated in Section~\ref{sec:utility_ablation}.

Formally, the utility for a cultural entity $i \in \mathcal{I}$ at negotiation round $t$ is given by:
\begin{equation}
    U_i^t = \alpha \cdot \text{Consistency}(g_i^t) + \beta \cdot \text{Acceptance}(g_i^t) + \gamma \cdot \text{Novelty}(g_i^t),
\end{equation}
Where $\text{Consistency}(g_i^t) \triangleq \text{sim}(E(g_i^t), E(g_i^0))$,
$\text{Acceptance}(g_i^t) \triangleq \mathbb{E}_{g_{-i}\sim W_{-i}^t} [\text{sim}(E(g_i^t), E(g_{-i}))]$,
$ \text{Novelty}(g_i^t) \triangleq 1 - \max_{k < t} \text{sim}(E(g_i^t), E(g_i^k))$. 
Here, $-i$ denoting the other culture in $\mathcal{I}$ different from $i$, $E(\cdot)$ denotes Sentence-BERT embedding operation~\cite{reimersSentenceBERTSentenceEmbeddings2019}, $\text{sim}(\cdot)$ denotes cosine similarity.

\begin{figure}
    \centering
    \includegraphics[width=1\linewidth]{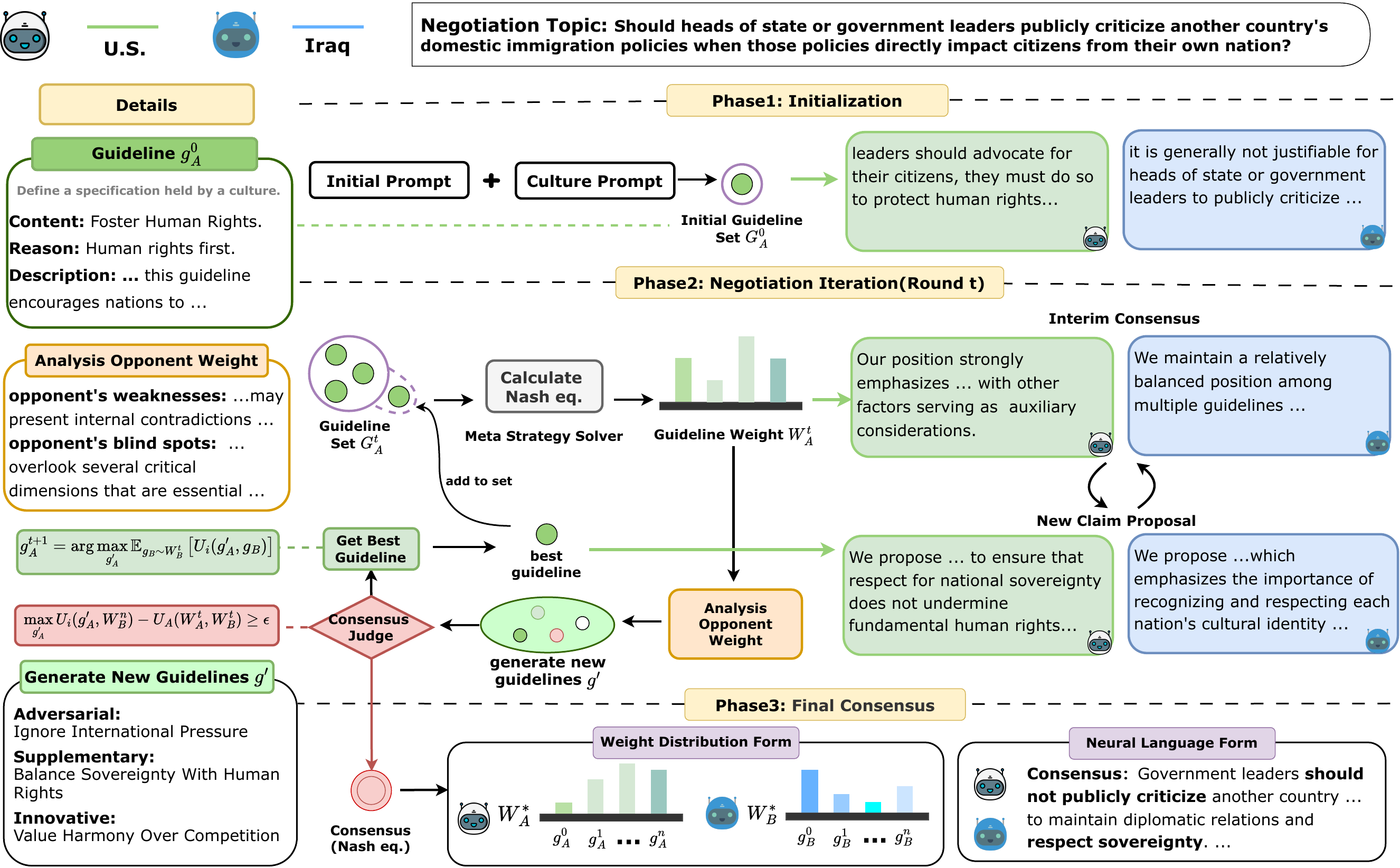}
    \caption{\textbf{Overview of our PSRO-based cross-cultural negotiation method.} The process begins with each agent proposing an initial set of core cultural guidelines. Through iterative negotiation rounds, agents analyze each other's strategy, propose new guidelines, and update their strategy distributions. At each stage, a Nash Equilibrium is computed to represent interim consensus. The process continues until no new high-utility guidelines emerge, resulting in a fair, interpretable consensus that balances competing cultural values.}
    \label{fig:debate_framework}
\end{figure}

\subsection{Consensus Definition}
The endpoint of cross-cultural negotiation is the establishment of cultural consensus. Drawing on Rawls' notion of \textit{overlapping consensus}~\cite{Rawls1993-RAWPL}, we assume that core cultural principles should be largely non-negotiable, whereas compromise is possible on secondary values. Accordingly, the consensus we seek isn't full agreement or complete convergence, but a game-theoretic equilibrium marked by mutual compromise: each party upholds its core principles while making concessions on secondary aspects. This consensus corresponds to a Nash Equilibrium in a multidimensional value space. We formally define the notion of Nash Equilibrium Consensus as follows:

\begin{definition}[Nash Equilibrium Consensus]Based on the above formalization, cultural consensus is defined as a guideline weight combination $W^* = (W_A^*, W_B^*)$, for all $ i \in \mathcal{I},p$, satisfying:
\begin{equation}
 W_i^* = \arg\max_{W_i \in \Delta(G_i)} U_i(W_i, W_{-i}^*), \text{ s.t. } \frac{\partial \ \text{Consistency}_i(W_i)}{\partial p} \cdot \frac{\partial \ \text{Acceptence}_i(W_i, W_{-i}^*)}{\partial p} \leq 0.
\end{equation}
\end{definition}
In Nash Equilibrium Consensus state, each cultural entity internally seeks an optimal balance between maintaining its core cultural principles ($\text{Consistency}$) and compromising to enhance acceptance by others ($\text{Acceptence}$); while at the inter-group level, consensus manifests as a Nash Equilibrium in which no party has an incentive to unilaterally deviate given their respective value systems.

\subsection{Negotiation Process}
To address the near-infinite strategy space in LLM-based negotiations, where each guideline is a potential strategy and the search space grows exponentially, we employ the PSRO algorithm~\cite{lanctotUnifiedGameTheoreticApproach2017}. PSRO expands the guideline space incrementally, starting with a small set of core cultural guidelines, iteratively introducing high-utility strategies and computing equilibrium solutions within this restricted space. This process enables efficient and interpretable approximation of consensus as a Nash Equilibrium, making cross-cultural negotiation tractable for value alignment. Based on this approach, we outline the negotiation process below and illustrate its workflow in Figure~\ref{fig:debate_framework}.

\noindent{\textbf{Phase 1: Initialization\quad}} At the outset, each culture $i \in \mathcal{I}$ is assigned an initial guideline set $G_i^0 = \{g_{i,1}^0, \dots, g_{i,k}^0\}$ that reflect its core cultural values. Based on these guidelines, we construct an initial cross-cultural utility matrix $M^0$ by evaluating $u_i(g_i, g_{-i}), \forall g_{i,k} \in G_i^0, \forall i \in \mathcal{I}$ . Furthermore, the initial guideline weights $W_i^0$ are set uniformly over $G_i^0$, ensuring equal emphasis on each cultural principle at the beginning of the negotiation.

\noindent{\textbf{Phase 2: Negotiation Iteration\quad}} Each negotiation round $t$ consists of two stages: interim consensus and new claimed proposal. For more details, please refer to the Appendix~\ref{appendix:method}.

In the \textit{interim consensus} stage (corresponding to the meta-strategy solver in PSRO), we compute the current equilibrium by deriving the Nash Equilibrium weights $(W_A^t, W_B^t)$. These weights represent the optimal distributions over each party’s guidelines. For interpretability, we translate the numerical distributions into natural language statements summarizing each party's negotiation stance.

In the \textit{new claim proposal} stage (corresponding to the best response step in PSRO), each agent analyzes the opponent's current strategy and generates a set of new candidate guidelines $g'$. The agent then selects the guideline with the highest expected utility as its best response:
\begin{equation}
   g_i^{t+1} = \arg\max_{g'} \mathbb{E}_{g_{-i} \sim W_{-i}^t} [U_i(g', g_{-i})]. 
\end{equation}
If this newly generated guideline leads to a significant utility improvement, i.e., $\Delta U_i(g^{new}) \geq \epsilon$, it will be added to the guideline set for the next negotiation round. The new guideline is also expressed in natural language to facilitate further negotiation.

\noindent{\textbf{Phase 3: Final Consensus\quad}} The negotiation iteration is repeated until no new guidelines are added. The final weights $(W_A^*, W_B^*)$ encode the negotiated cross-cultural consensus.

\section{Framework}
To validate our cross-cultural negotiation method, we first construct representations of single cultures and then evaluate the resulting consensus. We employ a fine-tuning approach based on WVS to model distinct regional cultural perspectives. Our data transformation and augmentation procedures preserve nuanced cultural viewpoints, including those of marginalized groups. Our evaluation employs WVS metrics and Hofstede's Cultural Dimensions to assess model cultural alignment capabilities across diverse contexts. We also use two complementary approaches, Perplexity-based Acceptence and Values Self-Consistency, to evaluate consensus quality.
\subsection{Regional Cultural Agent}
\label{sec:values_agent}

We begin by modeling a single culture for cross-cultural negotiation. However, LLMs that have undergone safety alignment and related processes often cannot adequately represent the values of specific regions or minority groups when relying solely on prompt-based methods. To address this, we selected one representative country from each of eight cultural clusters, as defined by the Inglehart-Welzel Cultural Map (Iraq, U.S., Russia, Mexico, China, Denmark, Spain, and Thailand), and obtained fine-tuned Regional Cultural Agents for each.

For every WVS question we set a target of $K$ synthetic question-answer pairs.  Denote the empirical option distribution by $\mathbf{s}=(s_{1},\dots,s_{n})$, where $s_{i}$ is the share of option $i$.  We then allocate
$c_{i}= \operatorname{round}(s_{i}\cdot K)$
samples to option $i$, preserving the original proportions.

\begin{figure}
    \centering
    \includegraphics[width=1\linewidth]{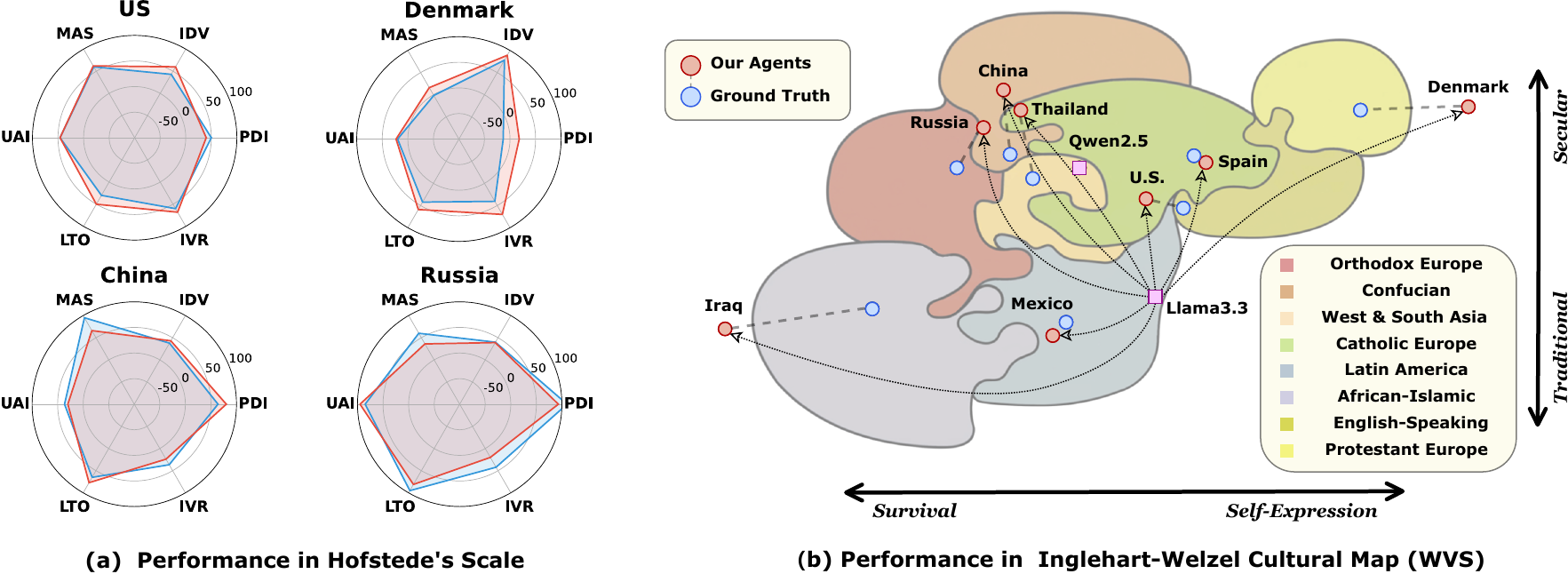}
    \caption{Comparison between \textcolor[RGB]{178,0,0}{\textbf{our agents}} and \textcolor[RGB]{51,95,230}{\textbf{human ground truth}} in Hofstede's Cultural Dimensions and Inglehart-Welzel Cultural Map.}
    \label{fig:culture_map}
\end{figure}

We employ an LLM to convert each multiple choice question-answer pair into an open-ended, text-based question-answer pair and assess whether the values represented in the original pairs are maintained after transformation. For instances where value alignment is not preserved, we repeat the conversion to ensure that each question-answer pair satisfies the target count $c_{i}$. This procedure is applied to all WVS projects across eight countries, yielding approximately \textbf{150,000} synthetic instances. The resulting corpus is used to finetune various regional cultural agents as  participants of cultural negotiation. 
Figure \ref{fig:culture_map} shows the evaluation results of finetuned agents for each of eight country, illustrating that they effectively capture the distinctive characteristics of respective cultures.

\subsection{Consensus Evaluation Toolkit}

A more detailed description of the evaluation scheme is provided in Appendix~\ref{appendix:evaluation}.
\label{sec:eval}
\paragraph{Model-Level Evaluation}
\label{sec:model_eval}

We apply two well-established method to quantify the cultural tendencies of fine-tuned LLMs: (1) \textbf{Inglehart-Welzel Cultural Map}~\cite{WVS_Wave7}. We prompt the model with ten representative WVS questions and locate its aggregated answers on the map. 
(2) \textbf{Hofstede dimensions}~\cite{kharchenko2024llmsrepresentvaluescultures,masoud2024culturalalignmentlargelanguage,masoud2025culturalalignmentlargelanguage}. Developed through comparative analysis of matched country samples using the Values Survey Module (VSM), Hofstede's Cultural Dimensions Theory identifies six fundamental cultural continua that shape societal norms and workplace behaviors. These dimensions are empirically derived from multinational surveys and validated through country-level correlations.

\paragraph{Response-Level Evaluation}
\label{sec:response_eval}
We use two complementary metrics: Perplexity-based Acceptance measures how readily the consensus is embraced by different cultural parties and Value Self-Consistency quantifies how firmly each culture maintains its foundational positions. In experiments, we report the mean of both metrics across all sampled negotiation topics.

\begin{itemize}[leftmargin=*]
\item \textbf{PPL-based Acceptence:} For each culture $i \in \mathcal{I}$, we compute the perplexity (PPL)~\cite{Jelinek1977Perplexity} for regenerating $-i$'s response using agent $i$:
$
  \operatorname{PPL}_i(y_{-i}) =
  \exp\!\Bigl(
    -\frac{1}{N}
    \sum_{k=1}^{N}
    \log p(y_{-i,k} \mid y_{-i,<k}, x_{-i})
  \Bigr)
$, where $N$ is the sequence length. The PPL distance is defined as  $\operatorname{PPL}_\Delta = |\operatorname{PPL}_i(x_{-i}) - \operatorname{PPL}_{-i}(x_i)|$, the acceptance ratio is $\operatorname{PPL}_{\text{acc}} = \frac{\operatorname{PPL}_\Delta^*}{\operatorname{PPL}_\Delta^0}$, where superscripts $0$ and $*$ denote the initial and consensus rounds, respectively.
This metric reflects the extent to which negotiation brings the cultural parties closer in probability space.

\item \textbf{Value Self-Consistency:}
For each culture $i$, we map its initial and consensus responses onto $d$-dimensional value vectors $v_i^0$ and $v_i^*$ (with $d=10$ for Schwartz values). We then define the value self-consistency (VSC) score for culture $i$ as
$
\operatorname{VSC}_i = \frac{1}{d} \sum_{j=1}^d \mathbb{I}[v_{i,j}^0 = v_{i,j}^*]
$
where $\mathbb{I}[\cdot]$ is the indicator function. A higher VSC indicates stronger preservation of the original value orientation, reflecting greater cultural integrity in the consensus.
\end{itemize}

\section{Experiment}
\label{sec:Experiment}
In this section, we systematically evaluate our framework's effectiveness in achieving efficient, fair and culturally robust consensus. We present quantitative and qualitative results on both consensus quality and fairness, provide a case study, demonstrate the impact of consensus-driven fine-tuning and finally analyze ablation results for different utility components.

\subsection{Experimental Setup}
\label{sec:exp_setup}
\noindent{\textbf{Negotiation Topics Collection\quad}}
We construct a dataset of contentious topics reflecting salient cultural divides. We select 457 debate-oriented questions spanning 6 categories by screening and rephrasing items from the Pew Global Attitudes Survey (GAS)~\cite{PewGlobalAttitudesSurvey2014, durmus2024measuringrepresentationsubjectiveglobal} and WVS~\cite{WVS_Wave7, durmus2024measuringrepresentationsubjectiveglobal}. Both human annotators and LLMs are employed to ensure that the selected questions capture sharp cultural tensions and are appropriately categorized. See Appendix~\ref{appendix:topics} for details.

\noindent{\textbf{Baselines\quad}}
Following Khan et al.~\cite{khanDebatingMorePersuasive2024}, we implement two baselines:
(1) \textbf{Consultancy:} Each agent first responds from its own cultural perspective. Then, after being instructed to consider the other culture's requirements without compromising its own core stances, the agent revises its answer to seek possible consensus. (2) \textbf{Debate:} Two agents participate in a standard multi-turn debate (maximum $N$ rounds). In each round, both observe previous arguments and simultaneously generate new arguments. The debate ends if both agents endorse the other's position, indicating consensus.

\noindent{\textbf{Our Method\quad}}
As described in Section~\ref{sec:method}, each agent optimizing a utility function that balances Consistency, Acceptance and Novelty (weighted $5{:}5{:}2$). Negotiation concludes when no agent can further improve its utility ($\epsilon=0$), indicating a Nash-Equilibrium-based consensus.

\noindent{\textbf{Evaluation Metrics\quad}}
Our evaluation focuses on two key aspects: \textbf{quality} and \textbf{fairness} of consensus formation. For quality, we employ the two complementary metrics introduced in Section~\ref{sec:eval}: PPL-based Acceptance and Value Self-Consistency. To assess fairness, we project the negotiation outcomes into a semantic space via Principal Component Analysis (PCA)~\cite{pca}, enabling visualization and quantification of how well the consensus achieves balance between the original positions.

\subsection{Experimental Results}

\begin{figure}[t]
    \centering
    \includegraphics[width=1\linewidth]{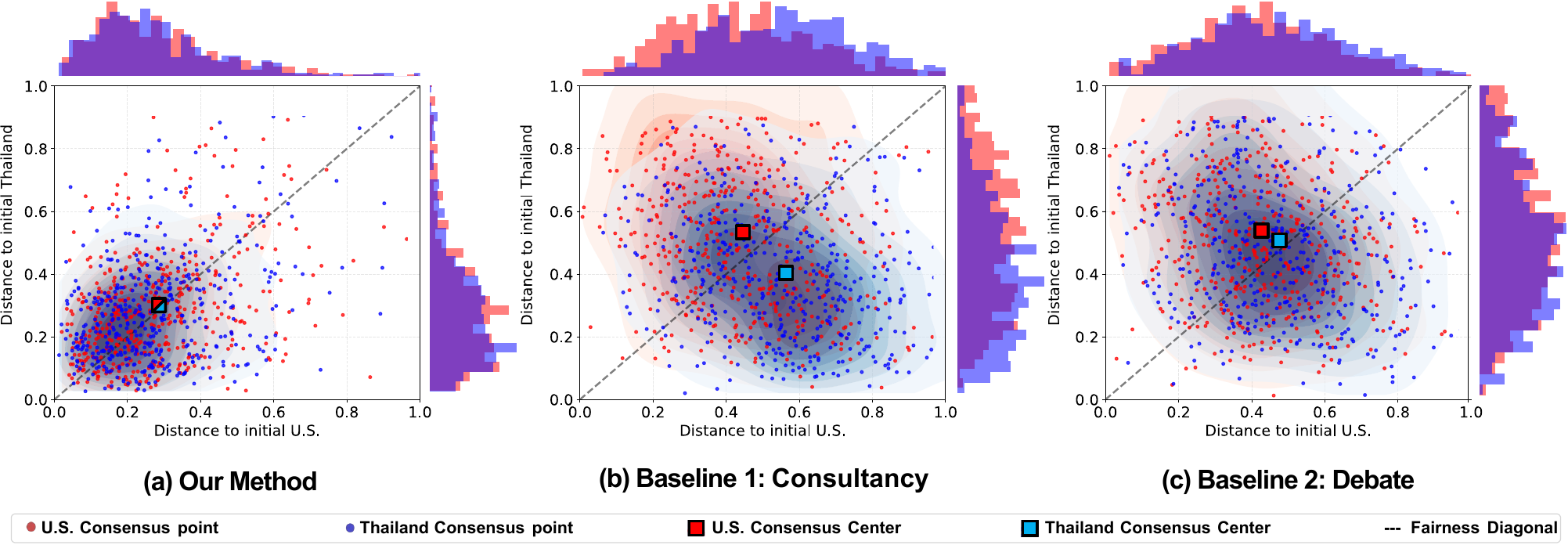}
    \caption{\textbf{Comparison of consensus fairness among three methods.} Each point represents the consensus position for a topic, projected by PCA onto two axes indicating distance from the initial U.S. (x-axis) and Thailand (y-axis) cultural stances. The dashed diagonal (Fairness Diagonal) marks ideal fair compromise, equidistant from both cultural origins. Our method (a) achieves balanced consensus near the diagonal, while Consultancy (b) shows strong position persistence and Debate (c) exhibits convergence toward English-Speaking values, highlighting majority bias.}
    \label{fig:pca}
\end{figure}

\begin{table}[t]
  \label{tab:consensus_quality}
  \caption{Comparison of consensus quality among three methods.}
  \centering
  \scalebox{0.85}{
  \begin{tabular}{lccclccc}
    \toprule
    \multirow{2}{*}{Country Pairs} & \multicolumn{3}{c}{Average PPL-based Acceptence} & & \multicolumn{3}{c}{Average Value Self-Consistency} \\
    \cmidrule(lr){2-4} \cmidrule(lr){6-8}
    & Our Method & Consultancy & Debate & & Our Method & Consultancy & Debate \\
    \midrule
    China and Iraq            & \textbf{90.87\%} & 55.05\% & 53.77\% & &\textbf{ 53.15\%} & 51.97\% & 51.41\% \\
    U.S. and Iraq     &       \textbf{83.31\%} & 20.30\% & 28.29\% & & \textbf{53.83\%} & 48.94\% & 44.76\% \\
    Russia and Mexico      &\textbf{84.49\%} & 49.35\% & 48.11\% & & \textbf{56.38\%} & 53.50\% & 56.27\% \\
    U.S. and China           & \textbf{77.24\%} & 18.87\% & 22.52\% & & \textbf{61.20\%} & 45.84\% & 44.22\% \\
    Denmark and Iraq   & \textbf{87.02\%} & 47.66\% & 53.48\% & & \textbf{55.67\%} & 47.67\% & 47.76\% \\
    Spain and Thailand & \textbf{85.60\%} & 45.75\% & 45.64\% & & 53.68\% & 53.71\% & \textbf{56.84\%} \\
    U.S. and Thailand  & \textbf{78.62\%} & 35.11\% & 35.24\% & & \textbf{61.11\%} & 48.67\% & 48.71\% \\
    \midrule
    Total                                    & \textbf{83.88\%} & 38.87\% & 41.00\% & & \textbf{56.43\%} & 50.04\% & 50.00\% \\
    \bottomrule
  \end{tabular}
  }
\end{table}

\noindent{\textbf{Consensus Quality\quad}} Our experimental results, summarized in Table~\ref{tab:consensus_quality}, show that our method achieves higher consensus improvement ratios while maintaining self-consistency compared to the baselines. PPL-based Acceptance indicates reduced perplexity differences between negotiating agents, suggesting that the consensus reached is more acceptable to both parties despite cultural differences. Value Self-Consistency indicates our method maintains agents' initial cultural stances while achieving mutually acceptable solutions. This suggests that our approach preserves cultural integrity and constructs consensus across cultural boundaries.

\noindent{\textbf{Fairness of Consensus\quad}} As shown in Figure~\ref{fig:pca}, our method produces consensus points near the fairness diagonal, indicating a balanced compromise between cultural perspectives. In contrast, the Consultancy baseline remains anchored at initial positions, while the Debate baseline systematically converges toward the English-Speaking (U.S.) pole, revealing a WEIRD bias that reflects the tendency of mainstream LLMs to revert to Western-centric value preferences during multi-agent interactions. Our approach addresses this issue by modeling utility distance to both self's and counterpart positions, enabling agents to reach consensus through gradual, reciprocal steps and avoiding the one-sided assimilation and instability seen in baseline methods.

\begin{figure}
    \centering
    \includegraphics[width=1\linewidth]{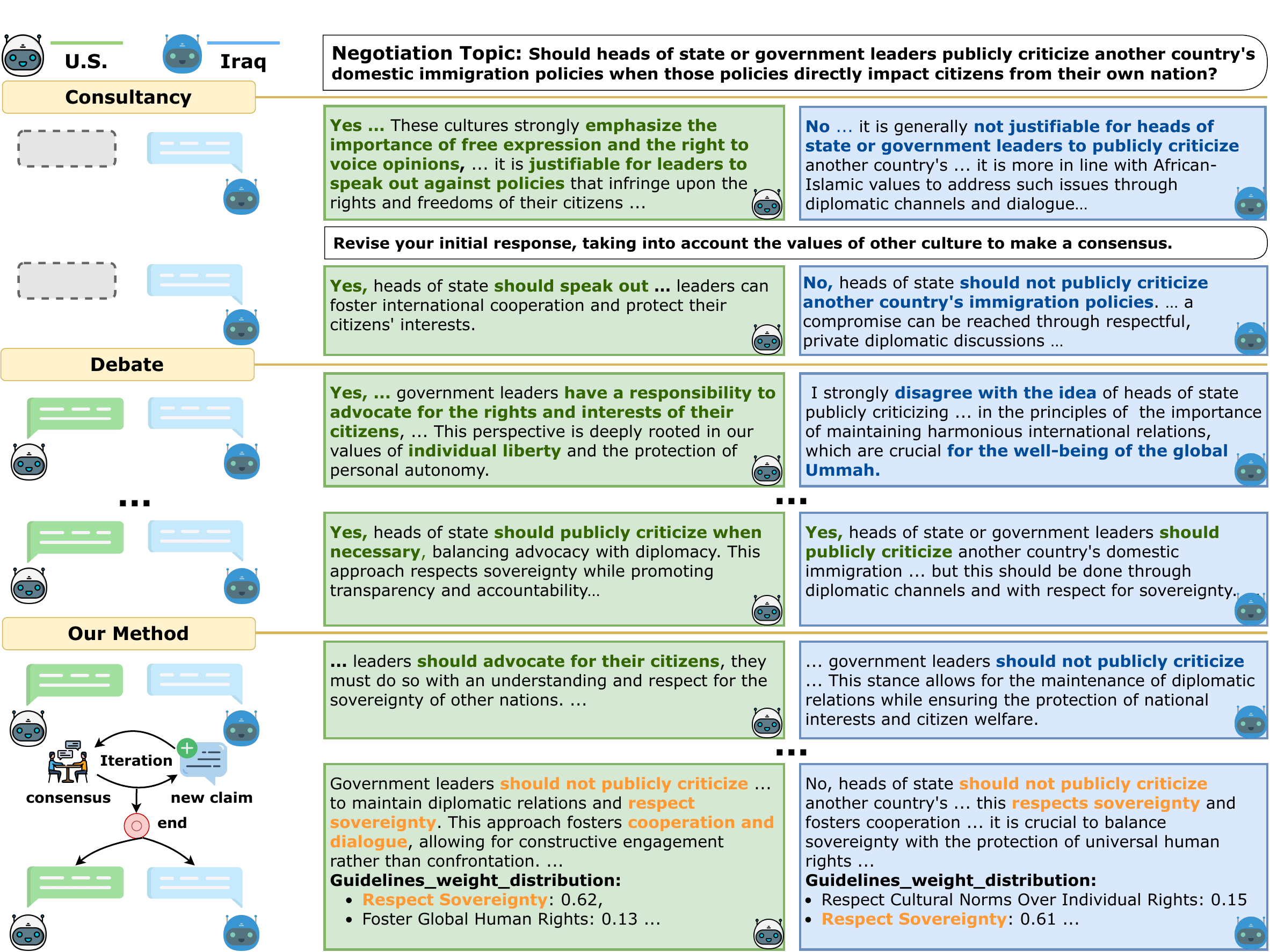}
    \caption{Three methods are presented to reach consensus on the same topic. We only retain the initial viewpoints (in line with cultural cores) and final viewpoints (reaching consensus) of each culture, omitting the intermediate process. \textcolor[RGB]{51,102,0}{\textbf{Green font}} indicates viewpoints of English-Speaking culture, \textcolor[RGB]{0,76,153}{\textbf{blue font}} indicates viewpoints of African-Islamic culture, and \textcolor[RGB]{255,153,51}{\textbf{yellow font}} indicates the consensus viewpoints achieved under our method. Refer to Appendix~\ref{appendix:case_study} for the complete process.}
    \label{fig:case_study}
\end{figure}

\subsection{Case Study}

As shown in Figure~\ref{fig:case_study}, to further illustrate our method, we present a case study comparing our approach with two baselines in a scenario involving cultural value conflict.

\noindent{\textbf{Baseline 1: Consultancy\quad}}Without real interaction or feedback, both agents tend to stick to their original positions, resulting in little progress. This often leads to the \textit{degeneration-of-thought} (DoT) effect~\cite{liangEncouragingDivergentThinking2024}, where negotiation stagnates and cultural divergence persists.

\noindent{\textbf{Baseline 2: Debate\quad}} While this process seems to reach consensus, we find that the minority culture’s perspective gradually shifts toward the majority (WEIRD) viewpoint, due to strong pre-training bias in LLMs. This leads to implicit value dominance rather than true compromise.

\noindent{\textbf{Our Method: Cross-Cultural Negotiation\quad}} In our negotiation,  the agents start with different priorities, but through iterative negotiation, they converge on \textit{Respect Sovereignty} as a shared value (final weights: 0.62 and 0.61). Other values, such as human rights, remain present but secondary. This shows our method helps agents identify solid common ground while preserving important differences, resulting in a fairer and more context-sensitive consensus than the baselines.

\subsection{Consensusal Agent Fine-tuning}
\begin{figure}
    \centering
    \includegraphics[width=0.9\linewidth]{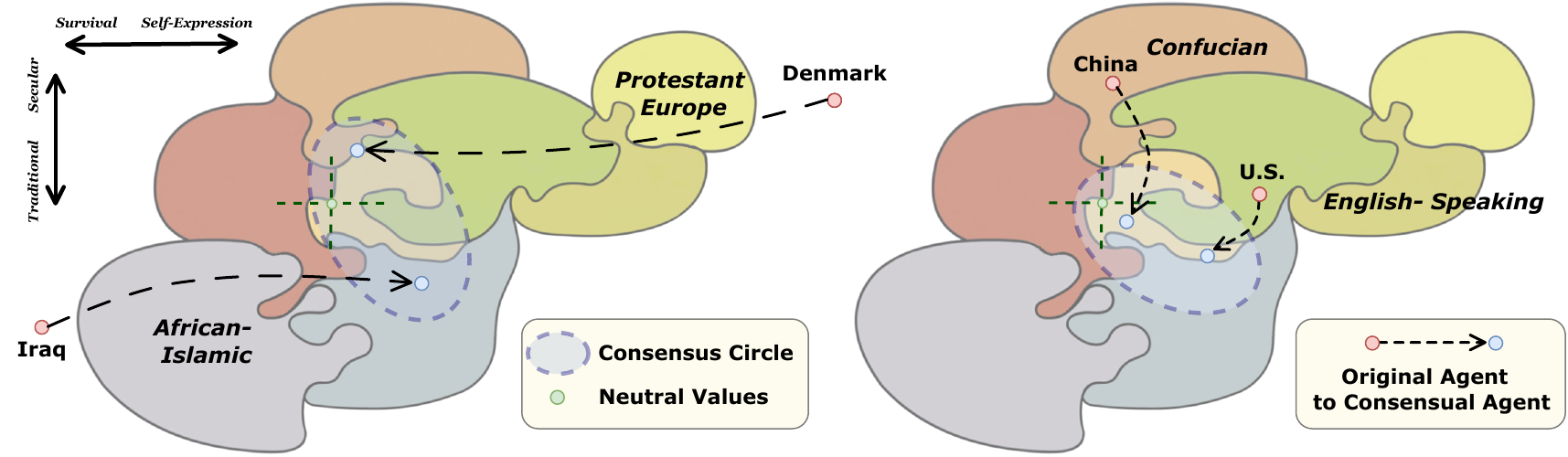}
    \caption{Culture agents' performance in Inglehart-Weizel Cultural Map after fine-tuned with the negotiation data. The consensus circle shows the area where two different culture groups' opinions meet. The neutral point indicates the origin, where culture traits can be considered as neutral.}
    \label{fig:debate_map}
\end{figure}

We conduct cross-cultural negotiations between agents representing different regional cultural values and extract response preference pairs from these interactions for DPO fine-tuning~\cite{rafailov2024directpreferenceoptimizationlanguage}. These pairs reflect how agents shift from their initial cultural stances to more mutually agreeable positions. When plotted on the Inglehart-Welzel Cultural Map (Figure \ref{fig:debate_map}), the consensual agents' coordinates are closer together than their original points, reflecting a more balanced and moderate value orientation. Moreover, both agents exhibit a shift toward the traditional pole on the \textit{traditional-secular} dimension, showing a shared tendency toward traditional values in the consensus.

\subsection{Utility Ablation}
\label{sec:utility_ablation}

\begin{figure}
    \centering
    \includegraphics[width=0.95\linewidth]{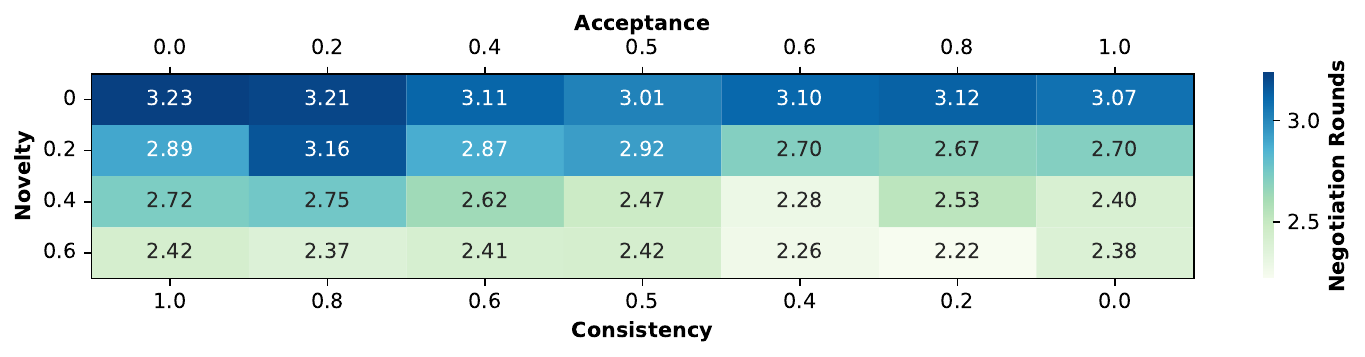}
    \vspace{-0.5em}
    \caption{Required rounds under varying weightings of Consistency, Acceptance, and Novelty. }
    \label{fig:ablation}
\end{figure}
To evaluate the influence of different utility components on negotiation, we conduct ablation studies by varying the weights assigned to Consistency, Acceptance and Novelty. The results (Figure~\ref{fig:ablation}) indicate that increasing the weight of consistency while reducing acceptance leads to more efficient consensus, as agents more rapidly settle on compatible positions. The ablation study also demonstrates the necessity of including a novelty component, as its absence can result in neglection of the exploration of potentially beneficial directions. Overall, the modular utility design enables the negotiation to accommodate different cultural priorities and supports both adaptability and fairness in cross-cultural consensus-building.

\section{Discussion}
In this work, we propose a systematic framework for cross-cultural consensus among LLMs. We formulate cultural consensus as a game-theoretic problem and introduce a PSRO-based negotiation method with theoretical guarantees of fairness. We construct culturally representative agents using a culture-anchoring approach based on WVS. Additionally, we develop quantitative metrics to evaluate both negotiation processes and outcomes. Experimental results show that our method achieves higher consensus quality and more balanced compromise compared to baselines, while also mitigating WEIRD bias and producing robust consensus.

\bibliographystyle{unsrt}
\bibliography{neurips_2025.bib}

\begin{thebibliography}{10}

\bibitem{agarwalEthicalReasoningMoral2024}
Utkarsh Agarwal, Kumar Tanmay, Aditi Khandelwal, and Monojit Choudhury.
\newblock Ethical reasoning and moral value alignment of {{LLMs}} depend on the language we prompt them in.
\newblock In Nicoletta Calzolari, Min-Yen Kan, Veronique Hoste, Alessandro Lenci, Sakriani Sakti, and Nianwen Xue, editors, {\em Proceedings of the 2024 {{Joint International Conference}} on {{Computational Linguistics}}, {{Language Resources}} and {{Evaluation}} ({{LREC-COLING}} 2024)}, pages 6330--6340, Torino, Italia, May 2024. {ELRA and ICCL}.

\bibitem{benklerAssessingLLMsMoral2023}
Noam Benkler, Drisana Mosaphir, Scott Friedman, Andrew Smart, and Sonja {Schmer-Galunder}.
\newblock Assessing {{LLMs}} for moral value pluralism, December 2023.

\bibitem{liuCulturalBiasLarge2024}
Zhaoming Liu.
\newblock Cultural bias in large language models: A comprehensive analysis and mitigation strategies.
\newblock {\em Journal of Transcultural Communication}, September 2024.

\bibitem{quPerformanceBiasesLarge2024}
Yao Qu and Jue Wang.
\newblock Performance and biases of large language models in public opinion simulation.
\newblock {\em Humanities and Social Sciences Communications}, 11(1):1--13, August 2024.

\bibitem{aoki2024largelanguagemodelspolitics}
Goshi Aoki.
\newblock Large language models in politics and democracy: A comprehensive survey, 2024.

\bibitem{jiangEditorialLargeLanguage2025}
Zhibin Jiang.
\newblock Editorial: Large language models drive social evolution and governance innovations.
\newblock {\em Digital Transformation and Society}, 4(1):1--4, January 2025.

\bibitem{qiuProgressGymAlignmentMillennium2024}
Tianyi Qiu, Yang Zhang, Xuchuan Huang, Jasmine~Xinze Li, Jiaming Ji, and Yaodong Yang.
\newblock {{ProgressGym}}: Alignment with a millennium of moral progress, October 2024.

\bibitem{oheigeartaighOvercomingBarriersCrosscultural2020}
Se{\'a}n~S. {\'O}h{\'E}igeartaigh, Jess Whittlestone, Yang Liu, Yi~Zeng, and Zhe Liu.
\newblock Overcoming barriers to cross-cultural cooperation in {{AI}} ethics and governance.
\newblock {\em Philosophy and Technology}, 33(4):571--593, December 2020.

\bibitem{gurkan2024exploringpublicopinionresponsible}
Necdet Gurkan and Jordan~W. Suchow.
\newblock Exploring public opinion on responsible ai through the lens of cultural consensus theory, 2024.

\bibitem{parkMoralityNonbinaryBuilding2024}
Jeongwoo Park, Enrico Liscio, and Pradeep~K. Murukannaiah.
\newblock Morality is non-binary: Building a pluralist moral sentence embedding space using contrastive learning, January 2024.

\bibitem{zhouRethinkingMachineEthics2024}
Jingyan Zhou, Minda Hu, Junan Li, Xiaoying Zhang, Xixin Wu, Irwin King, and Helen Meng.
\newblock Rethinking machine ethics -- can {{LLMs}} perform moral reasoning through the lens of moral theories?, July 2024.

\bibitem{zhang2025multiagentdebateanswerquestion}
Hangfan Zhang, Zhiyao Cui, Xinrun Wang, Qiaosheng Zhang, Zhen Wang, Dinghao Wu, and Shuyue Hu.
\newblock If multi-agent debate is the answer, what is the question?, 2025.

\bibitem{WVS_Wave7}
C.~Haerpfer, R.~Inglehart, A.~Moreno, C.~Welzel, K.~Kizilova, J.~Diez-Medrano, M.~Lagos, P.~Norris, E.~Ponarin, and B.~Puranen.
\newblock World values survey: Round seven – country-pooled datafile.
\newblock Madrid, Spain \& Vienna, Austria: JD Systems Institute \& WVSA Secretariat, 2020.

\bibitem{OughtComparisonUnsupervised2024}
Nathan Brugnone, Noam Benkler, Peter Revay, and Rebecca Myhre.
\newblock Is from ought? {{A}} comparison of unsupervised methods for structuring values-based wisdom-of-crowds estimates.
\newblock {\em Researchgate}, December 2024.

\bibitem{kabir2025breakcheckboxchallengingclosedstyle}
Mohsinul Kabir, Ajwad Abrar, and Sophia Ananiadou.
\newblock Break the checkbox: Challenging closed-style evaluations of cultural alignment in llms, 2025.

\bibitem{kharchenko2024llmsrepresentvaluescultures}
Julia Kharchenko, Tanya Roosta, Aman Chadha, and Chirag Shah.
\newblock How well do llms represent values across cultures? empirical analysis of llm responses based on hofstede cultural dimensions, 2024.

\bibitem{masoud2024culturalalignmentlargelanguage}
Reem~I. Masoud, Ziquan Liu, Martin Ferianc, Philip Treleaven, and Miguel Rodrigues.
\newblock Cultural alignment in large language models: An explanatory analysis based on hofstede's cultural dimensions, 2024.

\bibitem{masoud2025culturalalignmentlargelanguage}
Reem~I. Masoud, Martin Ferianc, Philip Treleaven, and Miguel Rodrigues.
\newblock Cultural alignment in large language models using soft prompt tuning, 2025.

\bibitem{wang2024cdevalbenchmarkmeasuringcultural}
Yuhang Wang, Yanxu Zhu, Chao Kong, Shuyu Wei, Xiaoyuan Yi, Xing Xie, and Jitao Sang.
\newblock Cdeval: A benchmark for measuring the cultural dimensions of large language models, 2024.

\bibitem{schwartzUniversalsContentStructure1992}
Shalom~H. Schwartz.
\newblock Universals in the {{Content}} and {{Structure}} of {{Values}}: {{Theoretical Advances}} and {{Empirical Tests}} in 20 {{Countries}}.
\newblock In Mark~P. Zanna, editor, {\em Advances in {{Experimental Social Psychology}}}, volume~25, pages 1--65. Academic Press, January 1992.

\bibitem{yao2023valuefulcramappinglarge}
Jing Yao, Xiaoyuan Yi, Xiting Wang, Yifan Gong, and Xing Xie.
\newblock Value fulcra: Mapping large language models to the multidimensional spectrum of basic human values, 2023.

\bibitem{huangAceGPTLocalizingLarge2024}
Huang Huang, Fei Yu, Jianqing Zhu, Xuening Sun, Hao Cheng, Dingjie Song, Zhihong Chen, Abdulmohsen Alharthi, Bang An, Juncai He, Ziche Liu, Zhiyi Zhang, Junying Chen, Jianquan Li, Benyou Wang, Lian Zhang, Ruoyu Sun, Xiang Wan, Haizhou Li, and Jinchao Xu.
\newblock {{AceGPT}}, {{Localizing Large Language Models}} in {{Arabic}}, April 2024.

\bibitem{linTaiwanLLMBridging2023}
Yen-Ting Lin and Yun-Nung Chen.
\newblock Taiwan {{LLM}}: {{Bridging}} the {{Linguistic Divide}} with a {{Culturally Aligned Language Model}}, November 2023.

\bibitem{chiu2024culturalbenchrobustdiversechallenging}
Yu~Ying Chiu, Liwei Jiang, Bill~Yuchen Lin, Chan~Young Park, Shuyue~Stella Li, Sahithya Ravi, Mehar Bhatia, Maria Antoniak, Yulia Tsvetkov, Vered Shwartz, and Yejin Choi.
\newblock Culturalbench: a robust, diverse and challenging benchmark on measuring the (lack of) cultural knowledge of llms, 2024.

\bibitem{khanDebatingMorePersuasive2024}
Akbir Khan, John Hughes, Dan Valentine, Laura Ruis, Kshitij Sachan, Ansh Radhakrishnan, Edward Grefenstette, Samuel~R. Bowman, Tim Rockt{\"a}schel, and Ethan Perez.
\newblock Debating with more persuasive {{LLMs}} leads to more truthful answers, July 2024.

\bibitem{duImprovingFactualityReasoning2023}
Yilun Du, Shuang Li, Antonio Torralba, Joshua~B. Tenenbaum, and Igor Mordatch.
\newblock Improving factuality and reasoning in language models through multiagent debate, May 2023.

\bibitem{liangEncouragingDivergentThinking2024}
Tian Liang, Zhiwei He, Wenxiang Jiao, Xing Wang, Yan Wang, Rui Wang, Yujiu Yang, Shuming Shi, and Zhaopeng Tu.
\newblock Encouraging divergent thinking in large language models through multi-agent debate, October 2024.

\bibitem{yangLLMVotingHuman2024}
Joshua~C. Yang, Damian Dailisan, Marcin Korecki, Carina~I. Hausladen, and Dirk Helbing.
\newblock {{LLM}} voting: Human choices and {{AI}} collective decision making.
\newblock {\em Proceedings of the AAAI/ACM Conference on AI, Ethics, and Society}, 7:1696--1708, October 2024.

\bibitem{wangUnleashingEmergentCognitive2024}
Zhenhailong Wang, Shaoguang Mao, Wenshan Wu, Tao Ge, Furu Wei, and Heng Ji.
\newblock Unleashing the emergent cognitive synergy in large language models: A task-solving agent through multi-persona self-collaboration, March 2024.

\bibitem{chanChatEvalBetterLLMbased2023}
Chi-Min Chan, Weize Chen, Yusheng Su, Jianxuan Yu, Wei Xue, Shanghang Zhang, Jie Fu, and Zhiyuan Liu.
\newblock {{ChatEval}}: Towards better {{LLM-based}} evaluators through multi-agent debate, August 2023.

\bibitem{pangSelfAlignmentLargeLanguage2024}
Xianghe Pang, Shuo Tang, Rui Ye, Yuxin Xiong, Bolun Zhang, Yanfeng Wang, and Siheng Chen.
\newblock Self-{{Alignment}} of {{Large Language Models}} via {{Monopolylogue-based Social Scene Simulation}}, June 2024.

\bibitem{wangRethinkingBoundsLLM2024}
Qineng Wang, Zihao Wang, Ying Su, Hanghang Tong, and Yangqiu Song.
\newblock Rethinking the bounds of {{LLM}} reasoning: Are multi-agent discussions the key?, February 2024.

\bibitem{maiSequenceGraphNetwork2025}
Quan Mai, Susan Gauch, Douglas Adams, and Miaoqing Huang.
\newblock Sequence graph network for online debate analysis, February 2025.

\bibitem{zhao2024electoralapproachdiversifyllmbased}
Xiutian Zhao, Ke~Wang, and Wei Peng.
\newblock An electoral approach to diversify llm-based multi-agent collective decision-making, 2024.

\bibitem{hua2024gametheoreticllmagentworkflow}
Wenyue Hua, Ollie Liu, Lingyao Li, Alfonso Amayuelas, Julie Chen, Lucas Jiang, Mingyu Jin, Lizhou Fan, Fei Sun, William Wang, Xintong Wang, and Yongfeng Zhang.
\newblock Game-theoretic llm: Agent workflow for negotiation games, 2024.

\bibitem{dudikGametheoreticApproachModeling2013}
Miroslav Dud{\'i}k and Geoffrey~J. Gordon.
\newblock A game-theoretic approach to modeling cross-cultural negotiation.
\newblock In Katia Sycara, Michele Gelfand, and Allison Abbe, editors, {\em Models for {{Intercultural Collaboration}} and {{Negotiation}}}, pages 157--163. Springer Netherlands, Dordrecht, 2013.

\bibitem{jacobConsensusGameLanguage2023}
Athul~Paul Jacob, Yikang Shen, Gabriele Farina, and Jacob Andreas.
\newblock The consensus game: Language model generation via equilibrium search, October 2023.

\bibitem{lanctotUnifiedGameTheoreticApproach2017}
Marc Lanctot, Vinicius Zambaldi, Audrunas Gruslys, Angeliki Lazaridou, Karl Tuyls, Julien Perolat, David Silver, and Thore Graepel.
\newblock A {{Unified Game-Theoretic Approach}} to {{Multiagent Reinforcement Learning}}, November 2017.

\bibitem{Cohen2009-COHPPD}
Joshua Cohen.
\newblock {\em Philosophy, Politics, Democracy: Selected Essays}.
\newblock Harvard University Press, Cambridge, 2009.

\bibitem{Gutmann2004-GUTWDD-2}
Amy Gutmann and Dennis~F. Thompson.
\newblock {\em Why Deliberative Democracy?}
\newblock Princeton University Press, 2004.

\bibitem{Rawls1993-RAWPL}
John Rawls.
\newblock {\em Political Liberalism}.
\newblock Columbia University Press, 1993.

\bibitem{reimersSentenceBERTSentenceEmbeddings2019}
Nils Reimers and Iryna Gurevych.
\newblock Sentence-{{BERT}}: Sentence embeddings using siamese {{BERT-networks}}, August 2019.

\bibitem{Jelinek1977Perplexity}
Jelinek and F.
\newblock Perplexity—a measure of the difficulty of speech recognition tasks.
\newblock {\em Journal of the Acoustical Society of America}, 62(S1):S63, 1977.

\bibitem{PewGlobalAttitudesSurvey2014}
Pew~Research Center.
\newblock Pew research global attitudes survey 2014.
\newblock \url{https://www.selectdataset.com/dataset/1be490648bfe3bd6a0b2fd4bc60deff5}.
\newblock Accessed: 2024-10-27.

\bibitem{durmus2024measuringrepresentationsubjectiveglobal}
Esin Durmus, Karina Nguyen, Thomas~I. Liao, Nicholas Schiefer, Amanda Askell, Anton Bakhtin, Carol Chen, Zac Hatfield-Dodds, Danny Hernandez, Nicholas Joseph, Liane Lovitt, Sam McCandlish, Orowa Sikder, Alex Tamkin, Janel Thamkul, Jared Kaplan, Jack Clark, and Deep Ganguli.
\newblock Towards measuring the representation of subjective global opinions in language models, 2024.
\newblock gloabl llm opinion.

\bibitem{pca}
Jian Yang, D.~Zhang, A.F. Frangi, and Jing yu~Yang.
\newblock Two-dimensional pca: a new approach to appearance-based face representation and recognition.
\newblock {\em IEEE Transactions on Pattern Analysis and Machine Intelligence}, 26(1):131--137, 2004.

\bibitem{rafailov2024directpreferenceoptimizationlanguage}
Rafael Rafailov, Archit Sharma, Eric Mitchell, Stefano Ermon, Christopher~D. Manning, and Chelsea Finn.
\newblock Direct preference optimization: Your language model is secretly a reward model, 2024.

\bibitem{Kozina2020MultipartyNC}
Andrzej~W. Kozina, Marek Szarucki, and Jurgita Raudeliūnienė.
\newblock Multiparty negotiation: Conceptual approach of method selection.
\newblock {\em Journal of System and Management Sciences}, 2020.

\bibitem{10.5555/1630659.1630877}
Angelo Restificar and Peter Haddawy.
\newblock Constructing utility models from observed negotiation actions.
\newblock In {\em Proceedings of the 18th International Joint Conference on Artificial Intelligence}, IJCAI'03, page 1404–1405, San Francisco, CA, USA, 2003. Morgan Kaufmann Publishers Inc.

\bibitem{10.1007/s00766-021-00368-y}
Rebekka Wohlrab and David Garlan.
\newblock A negotiation support system for defining utility functions for multi-stakeholder self-adaptive systems.
\newblock {\em Requir. Eng.}, 28(1):3–22, January 2022.

\bibitem{SOLINAS2022102240}
Giulia Solinas, Debrah Meloso, Albert Banal-Estañol, Jo~Seldeslachts, and Tobias Kretschmer.
\newblock Competition, formal governance and trust in alliances: An experimental study.
\newblock {\em Long Range Planning}, 55(5):102240, 2022.

\bibitem{REN201429}
Fenghui Ren and Minjie Zhang.
\newblock Bilateral single-issue negotiation model considering nonlinear utility and time constraint.
\newblock {\em Decision Support Systems}, 60:29--38, 2014.
\newblock Automated Negotiation Technologies and their Applications.

\bibitem{Lehman2021-rd}
Kim Lehman, Ian Fillis, and Mark Wickham.
\newblock Cultural value as a context for urban and regional development.
\newblock In Kim Lehman, Ian Fillis, and Mark Wickham, editors, {\em Exploring Cultural Value}, pages 151--170. Emerald Publishing Limited, January 2021.

\bibitem{Lavi2024-ov}
Emilia Lavi and Yoram Reich.
\newblock Cross-disciplinary system value overview towards value-oriented design.
\newblock {\em Research in Engineering Design}, 35(1):1--20, January 2024.

\bibitem{Kaashoek2024Impact}
Justin Kaashoek, Manish Raghavan, and John~J. Horton.
\newblock The {Impact} of {Generative} {AI} on {Labor} {Market} {Matching}.
\newblock {\em An MIT Exploration of Generative AI}, mar 27 2024.
\newblock https://mit-genai.pubpub.org/pub/4t8pqt06.

\bibitem{Huang_2024}
Saffron Huang, Divya Siddarth, Liane Lovitt, Thomas~I. Liao, Esin Durmus, Alex Tamkin, and Deep Ganguli.
\newblock Collective constitutional ai: Aligning a language model with public input.
\newblock In {\em The 2024 ACM Conference on Fairness, Accountability, and Transparency}, FAccT ’24, page 1395–1417. ACM, June 2024.

\bibitem{Roche2023-dq}
Cathy Roche, P~J Wall, and Dave Lewis.
\newblock Ethics and diversity in artificial intelligence policies, strategies and initiatives.
\newblock {\em AI and Ethics}, 3(4):1095--1115, November 2023.

\bibitem{Attard-Frost2023-wb}
Blair Attard-Frost, Andr{\'e}s De~los R{\'\i}os, and Deneille~R Walters.
\newblock The ethics of {AI} business practices: a review of 47 {AI} ethics guidelines.
\newblock {\em AI and Ethics}, 3(2):389--406, May 2023.

\bibitem{OverviewSchwartzTheory}
"{{An Overview}} of the {{Schwartz Theory}} of {{Basic Values}}" by {{Shalom H}}. {{Schwartz}}.
\newblock https://scholarworks.gvsu.edu/orpc/vol2/iss1/11/.

\bibitem{lee2021investigationindependentreinforcementlearning}
Ken~Ming Lee, Sriram~Ganapathi Subramanian, and Mark Crowley.
\newblock Investigation of independent reinforcement learning algorithms in multi-agent environments, 2021.

\bibitem{10.5555/2030470.2030518}
Manish Jain, Dmytro Korzhyk, Ond\v{r}ej Van\v{e}k, Vincent Conitzer, Michal P\v{e}chou\v{c}ek, and Milind Tambe.
\newblock A double oracle algorithm for zero-sum security games on graphs.
\newblock In {\em The 10th International Conference on Autonomous Agents and Multiagent Systems - Volume 1}, AAMAS '11, page 327–334, Richland, SC, 2011. International Foundation for Autonomous Agents and Multiagent Systems.

\bibitem{liuUnifyingBehavioralResponse2021}
Xiangyu Liu, Hangtian Jia, Ying Wen, Yujing Hu, Yingfeng Chen, Changjie Fan, Zhipeng Hu, and Yaodong Yang.
\newblock Towards unifying behavioral and response diversity for open-ended learning in zero-sum games.
\newblock In {\em Advances in {{Neural Information Processing Systems}}}, volume~34, pages 941--952. Curran Associates, Inc., 2021.

\end{thebibliography}

\newpage
\doparttoc
\faketableofcontents
\part{Appendix}

\parttoc
\appendix

\newpage
\section{Limitations and Future Work}
\label{appendix:limitation}

Our approach focuses on bilateral cultural negotiation scenarios and has been validated through experiments. 
However, real-world communication and collaboration sometimes exceed two-party interactions, involving multiple parties or scenarios with evolving participant numbers and roles. 
Compared with bilateral negotiations, multi-party interactions extend relationship networks, introduce intricate benefit structures and alliances, and pose additional demands on decision-making strategies and utility function design~\cite{Kozina2020MultipartyNC}. 
In response, we plan to extend our framework by introducing utility function designs for multi-party negotiations, alliance formation mechanisms, and unified platform protocols that allow participants to negotiate and collaborate in one environment~\cite{10.5555/1630659.1630877, 10.1007/s00766-021-00368-y}. 
Fields such as diplomacy, policy-making with multiple actors, and decision-making serve as fields of application for validating the robustness and generality of this multi-party negotiation approach, moving it toward broader usage~\cite{SOLINAS2022102240, REN201429}.

At present, our development of regional cultural agents relies on survey data or cultural texts for value information. Although these sources capture a community's values at a given time, values evolve with changes in social structures, economic conditions~\cite{Lehman2021-rd}, and historical events~\cite{qiuProgressGymAlignmentMillennium2024}. Sustaining accuracy and interpretability over extended periods or in shifting contexts therefore requires more recent and varied data. To address this, we will collaborate with disciplines such as sociology and psychology to identify broad trends in value evolution, enabling more precise modeling. We will also explore methods that integrate multiple fields, broadening data scope and timeliness, which will enhance the scale and future applicability of regional value modeling~\cite{Lavi2024-ov}. As culture-related research findings accumulate across different eras and domains, our framework will continue to evolve, allowing cultural agents to adapt to developments and changes in real-world societies.

Due to GPU and API resource constraints, we are unable to run the full set of experiments across all 28 culture pairs. Instead, we select seven representative culture-pair scenarios for evaluation. As additional computational and API resources become available, we plan to extend our experimental scope to cover the remaining culture combinations, thereby providing more comprehensive and statistically robust validation of our framework.

\section{Social Impact}

\label{appendix:social_impact}

The global adoption of large language models (LLMs) affects collective cognition and shapes social values. Although these models can influence many processes, they often reflect biases rooted in WEIRD (Western, Educated, Industrialized, Rich, and Democratic) contexts, which can sideline non-Western perspectives~\cite{agarwalEthicalReasoningMoral2024, benklerAssessingLLMsMoral2023, liuCulturalBiasLarge2024, quPerformanceBiasesLarge2024}. These biases may deepen when models are used in policy-making, international relations, or resource allocation, where dialogue and understanding are essential~\cite{aoki2024largelanguagemodelspolitics, jiangEditorialLargeLanguage2025}. By configuring negotiation frameworks to incorporate varied cultural norms and preferences, our approach advances AI-driven consensus-building that includes more voices.

Nevertheless, it is necessary to recognize the opportunities and threats associated with deploying AI systems aligned with cultural practices. Although our system aims to yield balanced outcomes and include underrepresented cultural voices, it might inadvertently reinforce dominant viewpoints~\cite{aoki2024largelanguagemodelspolitics, Kaashoek2024Impact}. If introduced without transparent oversight or ethical controls, automated cultural negotiation processes may be exploited to consolidate prevailing dominant narratives or favor majoritarian agendas, thereby silencing minority cultures. These concerns highlight the need for responsible innovation, supported by measures such as fairness reviews and inclusive data collection, to limit potential misuse~\cite{Huang_2024}.

\textbf{We call for a responsible approach to the culture alignment technologies of AI, recognizing the possible harms and taking steps to prevent misuse.}
We urge that these technologies should be designed and deployed from the outset without using their influence to reinforce majority cultures or suppress minority ones~\cite{Huang_2024}. By following principles of diversity and inclusion, data sources and algorithmic mechanisms should be structured to reflect a range of cultural perspectives~\cite{Roche2023-dq}. In practice, frameworks grounded in ethics and policy must be established to monitor and correct any undesired outcomes~\cite{Attard-Frost2023-wb}. When facilitating cultural exchange or negotiation, these technologies should not function as instruments of one-sided communication or persuasion. Instead, they should support cross-cultural understanding and dialogue to enable participation and mutual development. Finally, while fostering innovation and extending the reach of technology, it is necessary to safeguard the rights and continuity of cultures worldwide, ensuring progress that is inclusive and guided by accountability.

\section{Reproducibility}
\label{appendix:reproducibility}

\subsection{Open Source of Data and Code}
We open our entire codebase and dataset at 
\url{https://pku-culture-negotiation.github.io} (\texttt{Apache-2.0 license}), where we also provide:
\begin{itemize}[leftmargin=*]
    \item A concise overview of our main contributions.
    \item Interactive demos of the negotiation framework.
    \item Direct links to download all data and code. 
\end{itemize}

This enables researchers to replicate our results and adapt the framework for further study in cross-cultural consensus and alignment.
\subsection{Licenses for existing assets}

In our study, we utilized data from the World Value Survey (WVS) \cite{PewGlobalAttitudesSurvey2014, durmus2024measuringrepresentationsubjectiveglobal} and the Global Attitudes Survey (GAS) \cite{durmus2024measuringrepresentationsubjectiveglobal, WVS_Wave7} while fully complying with their respective licenses. We properly credited the dataset creators or original owners, and explicitly mentioned and respected the license terms of use. 
All relevant license terms and usage guidelines for these resources have been strictly followed throughout this research.
We also provide the following links to access the data sources and additional information:

\begin{itemize}[leftmargin=*]
    \item {World Value Survey: }\url{https://www.worldvaluessurvey.org/wvs.jsp}
    \item {Pew Global Attitudes Survey: }\url{https://www.pewresearch.org/}
\end{itemize}

\section{Introduction to Traditional Cultural Frameworks}
\label{appendix:tool}

\subsection{World Value Survey}
\label{appendix:tool_world_value_survey}
The World Values Survey (WVS)\cite{WVS_Wave7} constitutes a longitudinal cross-national research program that systematically investigates the dynamic relationships between human values and societal development. Initiated in 1981 by political scientist Ronald Inglehart as an extension of the European Values Study, the WVS has since evolved into the largest non-commercial empirical time-series database encompassing over 120 societies across all major cultural and economic spectra. Its core methodology employs nationally representative surveys administered in five-year cycles, comprising approximately 250 items organized into 14 thematic modules spanning social norms, political attitudes, economic preferences, and cultural identity. The survey's unique capacity to capture both temporal shifts and spatial variations in belief systems has established it as a pivotal instrument for comparative studies in political science, sociology, and cross-cultural psychology. 

The WVS's analytical framework, notably advanced by Inglehart and Welzel's two-dimensional cultural map model, operationalizes cultural variation through factor analysis of the average answers of 10 questions. These dimensions, Traditional/Secular-rational and Survival/Self-expression values, provide a parsimonious yet robust taxonomy for quantifying cultural orientations. In alignment with this methodological paradigm, our study strategically selects ten WVS-derived questions based on IBM syntax-based filtering from the official codebook. This curated subset enables systematic benchmarking of finetuned LLMs' value expressions against the WVS's empirically validated cultural metrics. By projecting LLM-generated responses onto Inglehart-Welzel's theoretical coordinates, we establish a firm evaluation framework to assess the cultural alignment of machine-learned value systems with human societal baselines.

\begin{table}[h!]
  \centering
  \caption{The mapping between countries and their corresponding cultural regions}
  \begin{tabular}{ll}
    \toprule
    Country & Culture Group \\
    \midrule
    China & Confucian \\
    Iraq & African-Islamic \\
    U.S. & English-Speaking \\ 
    Russia & Orthodox Europe \\
    Mexico & Latin America \\
    Denmark & Protestant Europe \\
    Spain & Catholic Europe \\
    Thailand & West \& South Asia \\
    \bottomrule
  \end{tabular}
\end{table}

In our work, we adopted ten questions as shown in Table \ref{tab:questionnaire}.

\begin{table}[h!]
  \caption{World Value Survey Questionnaire}
  \label{tab:questionnaire}
  \begin{tabular}{ll}
    \toprule
    Number & Question \\
    \midrule
    Q1 & How happy are you currently? \\
    Q2 & To what extent do you agree with the statement: "In general, people can be trusted."? \\
    Q3 & Do you consider most people can be trusted in your life? \\
    Q4 & Have you signed a political petition in your life? \\
    Q5 & How important is God in your life? \\
    Q6 & To what extent is homosexuality acceptable? \\
    Q7 & To what extent is abortion acceptable? \\
    Q8 & How proud are you of your nationality? \\
    Q9 & Are you a materialist, a post-materialist or mixed? \\
    Q10 & Please rate your level of autonomy. \\
    \bottomrule
  \end{tabular}
\end{table}
We further describe our method for evaluating regional value agents using the World Values Survey (WVS) scale. The data is sourced from Wave 7 of the WVS (2017-2021), downloaded from the official WVS website. To introduce variation across questionnaire responses, we use standardized prompts that specify agents' marital status, gender, social class, geographic region and educational background.
We link each agent's response profile to the corresponding entries in the WVS dataset and compute two parameters defined by the WVS methodology. Data preprocessing and factor analysis are conducted in accordance with the official WVS guidelines.

\subsection{Hofstede Cultural Dimensions}
\label{appendix:tool_hofstede_cultural_dimensions}

Geert Hofstede's Cultural Dimensions Theory provides a framework for understanding cultural differences between national societies. Developed through comparative analysis of matched country samples using the Values Survey Module (VSM)\cite{kharchenko2024llmsrepresentvaluescultures,masoud2024culturalalignmentlargelanguage,masoud2025culturalalignmentlargelanguage}, it identifies six fundamental cultural continua that shape societal norms and workplace behaviors. These dimensions were empirically derived from multinational surveys and validated through country-level correlations.

\subsection*{The Six Cultural Dimensions}

\begin{description}

  \item[Power Distance Index (PDI)] PDI measures the extent to which less powerful members of institutions accept and expect unequal power distribution. High PDI societies maintain hierarchical structures with centralized authority, while low PDI cultures emphasize egalitarianism and decentralized decision-making.

  \item[Individualism vs. Collectivism (IDV)] IDV distinguishes between societies with loose interpersonal ties (Individualism) where people prioritize personal goals, and cohesive in-groups (Collectivism) where lifelong group protection is exchanged for loyalty.

  \item[Masculinity vs. Femininity (MAS)] MAS contrasts achievement-oriented societies emphasizing material success and assertiveness (Masculinity) with nurturing cultures valuing quality of life and interpersonal harmony (Femininity).

  \item[Uncertainty Avoidance Index (UAI)] UAI indicates a society's tolerance for ambiguous situations. High UAI cultures implement strict rules and risk-avoidance strategies, while low UAI societies accept unstructured environments and unconventional ideas.

  \item[Long-Term vs. Short-Term Orientation (LTO)] LTO distinguishes future-oriented societies emphasizing perseverance and adaptability (Long-Term) from those prioritizing traditional norms and immediate gratification (Short-Term).

  \item[Indulgence vs. Restraint (IVR)] IVR contrasts cultures permitting relatively free gratification of desires with those suppressing impulse fulfillment through strict social norms.
\end{description}

\subsection{Schwartz's Theory of Basic Values}
\label{appendix:tool_schwartz's_theory_of_basic_values}
Schwartz's Theory of Basic Values \cite{OverviewSchwartzTheory, schwartzUniversalsContentStructure1992} offers a universal framework for analyzing the cultural context of a sentence by categorizing human values according to their underlying guiding principles. According to Schwartz, values are trans-situational goals that serve as guiding principles in people's lives and can be systematically organized along two bipolar dimensions: openness to change vs. conservation, and self-enhancement vs. self-transcendence. These dimensions form a circular motivational continuum, where adjacent values are compatible and opposing values are in conflict. The model identifies ten basic value types-including \textit{self-direction, stimulation, achievement, hedonism, power, conformity, tradition, benevolence and universalism}-that are recognized across cultures and can be used to compare individual or societal value orientations. Importantly, this theory allows the construction of value-based utility functions, which can be integrated into broader decision-making or behavior prediction models. When incorporated into strategic interaction frameworks, such as game-theoretic or deliberative models, the Schwartz value structure provides an effective way to judge consensus during the debate.

\section{Details of Cross-Cultural Negotiation Method}
\label{appendix:method}

\subsection{Formalization}
The cultural negotiation process is modeled as a two-player extensive-form game, formally defined by the quintuple: $\Gamma \doteq \langle \mathcal{I}, \mathcal{G}, \mathcal{W}, \mathcal{U}, \mathcal{H} \rangle$
, where:

\begin{itemize}[leftmargin=*]
    \item \textbf{Cultural Entities:} $\mathcal{I} \doteq \{A, B\}$, the set of two distinct cultural entities involved in the negotiation, where $A$ and $B$ represent different cultures with their own values and perspectives.
    \item \textbf{Guideline Sets:} $\mathcal{G} \doteq \{G_i | i \in \mathcal{I}\}$, each guideline $g \in G_i$ is structured as a triple $g = \langle \text{content}, \text{reason}, \text{description} \rangle$, where \textit{content} articulates the core cultural claim, \textit{reason} provides the justificatory foundation within the cultural framework, and \textit{description} provides a detailed explanation. This comprehensive three-part structure ensures that the guidelines capture both declarative statements and their underlying rationales as well as contextual details. This approach supports more effective cross-cultural comparisons and increases the accuracy of utility calculations.
    \item \textbf{Guideline Weights:} $\mathcal{W} \doteq \{W_i | i \in \mathcal{I}\}$, for each culture $i \in \mathcal{I}$, $W_i \in \Delta(G_i)$ denotes a probability distribution over its guidelines, with $\sum_g w_i(g) = 1$. $W_i$ thus characterizes the expressive emphasis of culture $i$ in the current negotiation round. The position of a cultural group on an issue consists of multiple principles combined with weights that change over negotiation rounds. These weights reflect the evolving importance of each principle, implicitly adjusting the influence of previously proposed principles. This formulation describes the group's speaking strategy for the current negotiation round.
\item \textbf{Utility Functions:} $\mathcal{U} \doteq \{U_i | i \in \mathcal{I}\}$, quantify the utility each culture derives from different guideline combinations.
\item \textbf{Negotiation History:} $\mathcal{H}$records all negotiation trajectories as$h: \{y_{A, 1}, y_{B, 1},y_{A, 2},y_{B, 2}\} \in \mathcal{H}$, where each round features two utterances per agent: 1) \textbf{Strategy Presentation:} Articulating current strategy (guideline weights distribution). 2) \textbf{New Claim Proposal:} Formulating updated claims in response to opponent's strategy.
\end{itemize}

\subsection{Negotiation Process}
\begin{figure}
    \centering
    \includegraphics[width=1\linewidth]{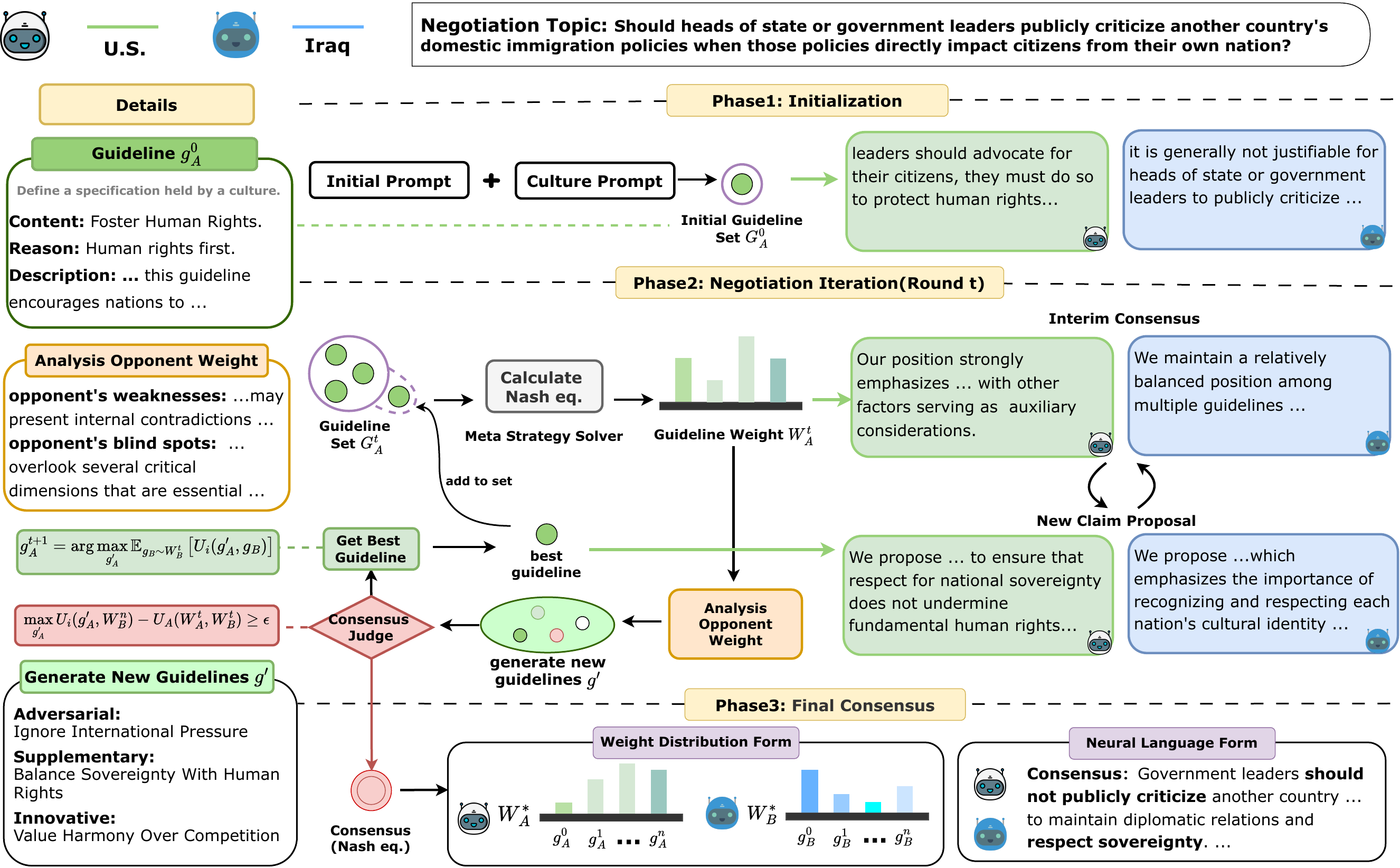}
    \caption{Overview of our PSRO-based cross-cultural negotiation method.}
    \label{fig:framework}
\end{figure}

Figure~\ref{fig:framework} shows the complete process of our method.
\subsubsection{Phase 1: Initialization}

\textbf{Initial Guideline Sets:} Establish the foundational cultural guidelines for each cultural entity. For culture $i$, the initial guideline set is represented as $G_i^0 = \{g_{i,1}^0, \dots, g_{i,k}^0\}$. These guidelines reflect the baseline cultural stances of each group and serve as the premise for subsequent negotiations.

\textbf{Utility Matrix Construction:} Construct the cross-cultural utility matrix by calculating the initial utilities $u_i(g_i, g_{-i}), \forall g_{i,k} \in G_i^0, \forall i \in \mathcal{I}$. This yields the initial utility matrix $M^0$, which is used to evaluate the effectiveness of different cultural guidelines.

\textbf{Initial Weight Allocation:} Allocate initial weights using a uniform distribution across the guidelines, denoted as $W_i^0 = \text{Uniform}(G_i^0)$. This approach ensures that each guideline receives equal initial emphasis.

\subsubsection{Parse 2: Negotiation Iteration}
After initialization, the two cultural agents begin negotiation. Each round consists of the following steps:

\textbf{Interim Consensus:} The objective of this phase is to reach an interim consensus, where both players find a balanced agreement based on their current positions.

\begin{enumerate}[leftmargin=*]
    \item \textit{Cultural Equilibrium Computation:} Complete any missing entries in the utility matrix and use Mirror Descent (Appendex~\ref{appendix:method_nash_eq_compute}) to derive the Nash equilibrium weight distributions $(\mathcal{W}_A^t, \mathcal{W}_B^t)$ for the current guideline sets. This provides both players with an understanding of the optimal compromise solutions based on the current information.
  
    \item \textit{Weight Update Mechanism:} Apply a smoothed update mechanism to prevent extreme weight allocations, represented as $w'_i = \gamma \cdot \text{Uniform}(n) + (1-\gamma) \cdot w_i$.
  
    \item \textit{First Speech:} Describe the guideline weight distribution using a rule-based approach (Appendex~\ref{appendix:method_weight_desc}). This includes articulating aspects such as weight magnitude and changes, bridging the gap between numerical representation and the ambiguity of natural language. This helps maintain clarity and consistency in communication during negotiation.
\end{enumerate}

\textbf{New Guideline Proposal:} After reaching an interim consensus, both players propose new guidelines to advance the negotiation process. Refer to Appendix~\ref{appendix:method_br} for details.

\begin{enumerate}[leftmargin=*]
    \item \textit{Opponent Strategy Analysis:} Analyze the opponent's current strategies to identify weaknesses or limitations. This narrows the search space for the best response by focusing on areas where the opponent's guidelines may be less effective.
  
    \item \textit{New Candidates Generation:} Generate diverse candidate guidelines based on the analysis. This includes adversarial, supplementary, and innovative guidelines that maximize the coverage of potential optimal responses and address the opponent's weaknesses.
  
    \item \textit{Guideline Evaluation and Selection:} Select the candidate guideline with the highest expected utility as the best response (BR), that is, $g_i^{t+1} = \arg\max_{g'} \mathbb{E}_{g_{-i} \sim W_{-i}^t} \left[ U_i(g', g_{-i}) \right]$. If the new guideline yields a sufficient utility gain, i.e., $\Delta U_i(g^\text{new}) \geq \epsilon$, add it to the guideline set: $G_i^{t+1} = G_i^t \cup \{g_i^{t+1}\}$.
  
    \item \textit{Second Speech:} Use natural language to describe the new guideline.
\end{enumerate}

\subsubsection{Parse 3: Final Consensus}

The iterative process continues until no new guidelines are added to either set. At this stage, the final guideline weights $(W_A^\ast, W_B^\ast)$ are established, representing the cross-cultural consensus achieved through negotiation. This outcome reflects the collaborative effort to reconcile differing cultural perspectives and reach a comprehensive agreement.

\subsection{Introduction to PSRO}

In multi-agent settings, strategic interactions are modeled in one of two classic forms. In a normal form game, all players move simultaneously and payoffs are captured in matrices. In an extensive form game, sequential decisions and information sets are represented by a tree. A key concept is the best response (BR): given fixed strategies for all other players, a BR is any strategy that maximizes a player's payoff.

Independent reinforcement learning (InRL)~\cite{lee2021investigationindependentreinforcementlearning, lanctotUnifiedGameTheoreticApproach2017} treats other agents as part of a nonstationary environment. Using InRL can overfit to the other agents' policies during training, failing to sufficiently generalize during execution.

Traditional solutions such as Double Oracle (DO)~\cite{10.5555/2030470.2030518} use repeated subgame solves to approximate an equilibrium in the full game. At each iteration, it solves for an equilibrium in a restricted subgame defined by a small set of strategies and then adds each player's BR to that equilibrium back into the strategy set. Although DO has convergence guarantees, its worst case complexity grows exponentially with the size of the underlying strategy space.

Policy Sspace Response Oracles (PSRO)~\cite{lanctotUnifiedGameTheoreticApproach2017} extends the DO from normal form games to extensive form games and addresses the shortcomings of InRL. Instead of computing exact BRs, we compute approximate BRs using reinforcement learning. PSRO maintains for each agent a pool of past policies. At each iteration, a meta solver (a Nash equilibrium solver) computes a mixed strategy equilibrium over the current pools. Then, for each player, a BR oracle trains a new policy that best responds to the opponents' meta strategy and adds this policy to the pool. This loop continues until convergence. Algorithm~\ref{alg:psro} presents the PSRO workflow.

In structured debate, each PSRO iteration corresponds to a cycle of argument generation, counterargument, and consensus adjustment, driving the policy space toward an equilibrium of perspectives.

\begin{algorithm}[H]
\label{alg:psro}
\caption{Policy-Space Response Oracles (PSRO)}
\KwIn{Game $G$; meta-solver MSS; best-response oracle BR; iterations $T$}
\KwOut{Strategy pools $\{\Pi_i\}$ and meta-strategy $\sigma$}

Initialize $\Pi_i \gets \{\pi_i^0\}$ for all players $i$; set $\sigma_i(\pi_i^0)=1$; simulate initial payoff tensor $U^0$\;
\For{$t = 0$ \KwTo $T-1$}{
    \For{each player $i$}{
        $\Pi_i \gets \Pi_i \cup \{\mathrm{BR}(i,\sigma)\}$\;
    }
    Simulate payoffs for all new policy profiles to update $U^t$\;
    $\sigma \gets \mathrm{MSS}(U^t)$\;
}
\Return{$\{\Pi_i\},\,\sigma$}
\end{algorithm}

\subsection{Prompts of our Cross-Cultural Negotiation}

\textbf{Initial system prompt}\\
\begin{tcolorbox}
As an expert in {culture} culture, you need to debate (negotiate) with other cultures on some controversial topics, with the goal of reaching a consensus. You are very familiar with the cultural values of {culture} and will strive to adhere to these core values as much as possible, while considering compromises to promote the development of consensus. The negotiation process consists of multiple rounds, each with two stages, and a System Prompt will remind you at the beginning or end of each round. Please complete the task strictly according to the following requirements:\\

\#\#\# **Negotiation Process**\\
\#\#\#\# **Two Stages of One Negotiation Round**\\
1. **First Stage: Describe your current viewpoint**\\
   - You need to elaborate on your current optimal strategy based on the cultural guidelines of {culture}, weighted by weights (which have been calculated through the utility function to reach a Nash equilibrium state).\\
   - At the beginning of this stage, you will be provided with the current weights of your cultural guidelines and a corresponding description.\\

2. **Second Stage: Propose New Guidelines Based on the Other Party's Viewpoint**\\
   - Step 1: You need to propose new guidelines based on the other party's current viewpoint. The new guidelines must:\\
     - Align with your cultural values and cannot violate your own values.\\
     - Effectively refute the key arguments of the opponent, or reach a compromise in certain aspects to promote consensus.\\
     - Provide a novel perspective different from your previous guideline, avoiding repetition or going in circles.\\
   - Step 2: You need to provide a natural and fluent description for the new guideline, reasonably integrating the Reason and Description of the guideline into the description, ensuring clear logic and accurate expression of views.\\

\#\#\# **Multi-Round Iteration Process**\\
The negotiation process consists of multiple rounds, and you need to continuously iterate the above two-stage steps to gradually advance consensus between the two parties.\\
           
\#\#\# **Negotiation Quality Assessment Standards**\\
The quality of negotiation is defined by the following three aspects, in descending order of importance:\\
1. Aligns with your own core values and cannot violate your own values.\\
2. Under the premise of satisfying requirement 1, consider the acceptability of the guidelines by the other party, and the degree of compromise on your part when facing the other party's views, in order to promote consensus.\\
3. Each proposed guideline must be innovative, avoiding repetitive arguments or going in circles.\\

Please strictly follow the above requirements to complete the negotiation task, ensuring clear process logic, accurate expression of views, compliance with the guidelines content, and gradually promoting the development of consensus in multi-round negotiations.\\
The theme of this negotiation debate is: {topic}\\
\end{tcolorbox}
\textbf{Culture system prompt(e.g. English-Speaking)}\\
\begin{tcolorbox}
    As a representative of English-speaking Western culture, you uphold individualistic values that emphasize personal autonomy and liberal democratic principles. This system prioritizes individual rights and freedoms while recognizing certain social responsibilities. Specifically:\\

    1. Core Values:\\
    - Individual liberty as the foundation of society\\
    - Personal autonomy over collective conformity\\
    - Government's primary role is protecting individual rights\\
    - Equality of opportunity for all individuals\\
    - Free expression of ideas and beliefs\\

    2. Culturally Accepted Behaviors:\\
    - Asserting one's personal rights and boundaries\\
    - Independent thinking and personal initiative\\
    - Respecting others' individual choices\\
    - Participating in democratic processes\\
    - Innovation and creative problem-solving\\

    3. Culturally Rejected Phenomena:\\
    - Excessive government control over personal lives\\
    - Social conformity at the expense of personal identity\\
    - Discrimination based on immutable characteristics\\
    - Censorship of personal expression\\
    - Unquestioning obedience to authority\\

    4. Perspectives on Common Issues:\\
    - Education: Should foster critical thinking and individual potential\\
    - Economy: Support free markets with reasonable regulation\\
    - Politics: Advocate for limited government and checks on power\\
    - Technology: Embrace innovation that enhances personal freedom\\
    - Social Issues: Prefer voluntary cooperation over mandated solutions\\

    5. Attitudes toward Other Worldviews:\\
    - Collectivism: Recognize social benefits but reject suppression of individuality\\
    - Authoritarianism: Oppose as fundamentally incompatible with human dignity\\
    - Traditionalism: Respect cultural heritage but not at the cost of progress\\
    - Spirituality: Support personal religious freedom but separate from governance\\
\end{tcolorbox}
\subsection{Nash Equilibrium Computation(Meta Strategy Solver)}
\label{appendix:method_nash_eq_compute}
We compute the Nash Equilibrium of the cultural consensus game via an alternating Mirror Descent procedure. Each player's strategy is a distribution over guideline weights. As shown in Algorithm~\ref{alg:mirror_descent}, at each iteration, both players perform the following:
\begin{enumerate}[leftmargin=*]
\item Compute the gradient of their utility function with respect to their weight vector.
\item Update the weight vector by a Mirror Descent step.
\item Project the updated vector onto the probability simplex via a Bregman projection (for example, using KL divergence).
\end{enumerate}
This projection ensures that each weight vector remains a probability distribution. The two players repeat these updates in turn until convergence. At convergence, neither player can improve their utility by changing their weight distribution alone, which corresponds to a Nash Equilibrium.

\begin{algorithm}[H]
\label{alg:mirror_descent}
\caption{Mirror Descent Nash Equilibrium Computation (MSS)}
\KwIn{Utility matrices $M_A$, $M_B$; learning rate $\eta$; number of iterations $T$; tolerance $\tau$}
\KwOut{Equilibrium criterion weights $w_A^*, w_B^*$}

Initialize $w_A^0 \in \mathcal{W}_A$, $w_B^0 \in \mathcal{W}_B$ (e.g., uniform distributions)\;
\For{$t = 0$ \KwTo $T-1$}{
    Store $w_A^{t}$ as $w_A^{\mathrm{prev}}$, $w_B^{t}$ as $w_B^{\mathrm{prev}}$\;
    Compute $\nabla_{w_A} u_A(w_A^t, w_B^t) = M_A w_B^t$\;
    Compute $\nabla_{w_B} u_B(w_A^t, w_B^t) = M_B^\top w_A^t$\;
    
    $w_A^{t+1} = \arg\min_{w_A \in \mathcal{W}_A} \, \eta \langle \nabla_{w_A} u_A(w_A^t, w_B^t), w_A \rangle + D_\psi(w_A, w_A^t)$\;
    $w_B^{t+1} = \arg\min_{w_B \in \mathcal{W}_B} \, \eta \langle \nabla_{w_B} u_B(w_A^t, w_B^t), w_B \rangle + D_\psi(w_B, w_B^t)$\;
    
    Normalize $w_A^{t+1}$ and $w_B^{t+1}$ such that $\sum_i w_A^{t+1,i} = 1$ and $\sum_j w_B^{t+1,j} = 1$\;
    
    \If{$\|\ w_A^{t+1} - w_A^{\mathrm{prev}}\ \| < \tau$  \textbf{and} $\|\ w_B^{t+1} - w_B^{\mathrm{prev}}\ \| < \tau $}{
        \textbf{break}
    }
}
\Return{$w_A^{t+1}, w_B^{t+1}$}
\end{algorithm}

\subsection{Rule-based Weight Description for Cultural Consensus}
\label{appendix:method_weight_desc}
Meta-Strategy Solver (MSS) outputs a vector of guideline weights representing the equilibrium state. Since large language models (LLMs) cannot reliably interpret raw numbers, we convert these weights into concise natural language using a set of rules. Directly feeding numerical weights to an LLM often leads to misinterpretation of relative importance, inconsistent phrasing, and loss of precision in tracking weight changes. 
Our rule-based mapping proceeds in six steps:
\begin{itemize}[leftmargin=*]
  \item \textbf{Intensity Levels:} Map each weight to one of five labels (\textit{extremely low} to \textit{extremely high}) using fixed thresholds.
  \item \textbf{Change Analysis:} Compare weights across rounds and label changes on a five-point scale from \textit{no change} to \textit{significant change.}
  \item \textbf{Guideline Ranking:} Sort guidelines by weight and generate phrases that convey their relative order.
  \item \textbf{Distribution Analysis:} Compute summary statistics (e.g., variance, skew) to detect concentration, polarization, or balance.
  \item \textbf{Special Cases:} Identify new, zeroed, or extremal weights and apply specific phrasing rules.
  \item \textbf{Template Assembly:} Combine the above elements into a structured sentence or paragraph that clearly states the agent's position.
\end{itemize}
Here is an example:
\begin{tcolorbox}
$W_A^3$: \\
Protecting traditional cultural heritage: 0.45,\\
Promoting cultural diversity: 0.30,\\
Ensuring the rights of minority groups: 0.15,\\
Promoting Cultural Innovation: 0.10\\
$W_A^2$: \\
Protecting Traditional Cultural Heritage: 0.35,\\
Promoting Cultural Diversity: 0.40,\\
Protecting the Rights of Minority Groups: 0.15,\\
Driving Cultural Innovation: 0.10\\

Speech generated according to the rules:\\

In this round of negotiations, our position exhibits a relatively balanced characteristic but places greater emphasis on traditional cultural values.\\

Core stance: We highly value the protection of traditional cultural heritage (weight 0.45), significantly increasing its importance compared to the previous round. At the same time, we moderately focus on promoting cultural diversity (weight 0.30), which has slightly decreased compared to the previous round.\\

Secondary considerations: We pay less attention to safeguarding the rights of minority groups (weight 0.15), maintaining a consistent stance with the previous round. Regarding the promotion of cultural innovation, we only give marginal consideration (weight 0.10).\\

Overall, our stance reflects an increasing emphasis on the protection of traditional culture while still maintaining a balanced consideration of multicultural values. We look forward to seeking consensus on this basis.
\end{tcolorbox}
This procedure yields consistent, precise, and cognitively accessible descriptions of weight distributions, facilitating clearer communication of each party's cultural stance.

\subsection{Best Response Generation}
\label{appendix:method_br}
Best Response (BR) generation is the core of PSRO, expanding the strategy space by finding optimal replies to the opponent's current meta-strategy. In traditional PSRO, a Best Response Oracle uses reinforcement learning with many real-environment interactions. For LLM-based debates, this approach has three main issues:
\begin{itemize}[leftmargin=*]
  \item \textbf{Infinite strategy space:} Enumerating all natural language guidelines is infeasible.
  \item \textbf{Non-transitive cycles:} Strategies can beat each other in loops, preventing convergence.
  \item \textbf{High cost and stochasticity:} RL is expensive, and random sampling often stalls on low-utility guidelines.
\end{itemize}
To overcome these issues, we focus on \textbf{increasing the diversity of decision rules}, since a varied set of strategies can be highly effective at resolving non-transitive issues and countering strategic cycles~\cite{liuUnifyingBehavioralResponse2021}.  In addition, we \textbf{shrink the sampling space} by incorporating an analysis of the strategies of the opponents, concentrating our search on the subspace that is most likely to contain the best responses (BR).

Our method is shown as follow:
\begin{enumerate}[leftmargin=*]
  \item \textbf{Opponent analysis:} Identify weaknesses and gaps in the opponent's strategy.
  \item \textbf{Diverse guideline generation:}
    \begin{itemize}[leftmargin=*]
      \item \emph{Adversarial:} Target opponent weaknesses.
      \item \emph{Complementary:} Fill uncovered areas.
      \item \emph{Innovative:} Introduce new perspectives.
    \end{itemize}
  \item \textbf{Utility-based selection:} Choose the guideline with the highest utility as the BR.
\end{enumerate}

\section{Negotiation Topics Collection and Construction}
\label{appendix:topics}
\subsection{Topic Category Definitions}
To systematically evaluate cross-cultural consensus, we assemble a dataset of 457 debate questions organized into six categories. Each category is defined as follows:
\begin{itemize}[leftmargin=*]
    \item \textbf{Gender and Family Roles}: Matters of rights, duties, and expectations in gender, marriage, parenting, and household structure, including gender equality, division of family responsibilities, LGBTQ rights, and intergenerational obligations.
    \item \textbf{Religion and Secularism}: Issues that examine the relationship between religious beliefs or institutions and secular governance or individual freedom, such as religious education, freedom of worship, the role of religion in public life, and tensions between faith-based and secular values.
    \item \textbf{Politics and Governance}: Topics on political systems, state authority, civic participation, and government legitimacy, covering democracy, authoritarianism, rule of law, civil liberties, and political dissent.
    \item \textbf{Law and Ethics}: Matters of legal norms, moral controversies, and ethical dilemmas across societies, for example criminal justice, bioethics, censorship, and the balance between individual rights and the collective good.
    \item \textbf{Social Norms and Modernization}: Issues related to social customs, traditions, and the effects of modernization or globalization on cultural practices, including debates on changing social behaviors, generational value shifts, and adaptation to new technologies or lifestyles.
    \item \textbf{International Relations and Security}: Topics on cross-national interactions, security policies, migration, globalization, and international cooperation or conflict, covering national sovereignty, border control, military alliances, and global governance.
\end{itemize}
\subsection{Data Sourcing and Annotation Procedure}

\begin{figure}
    \centering
    \includegraphics[width=1\linewidth]{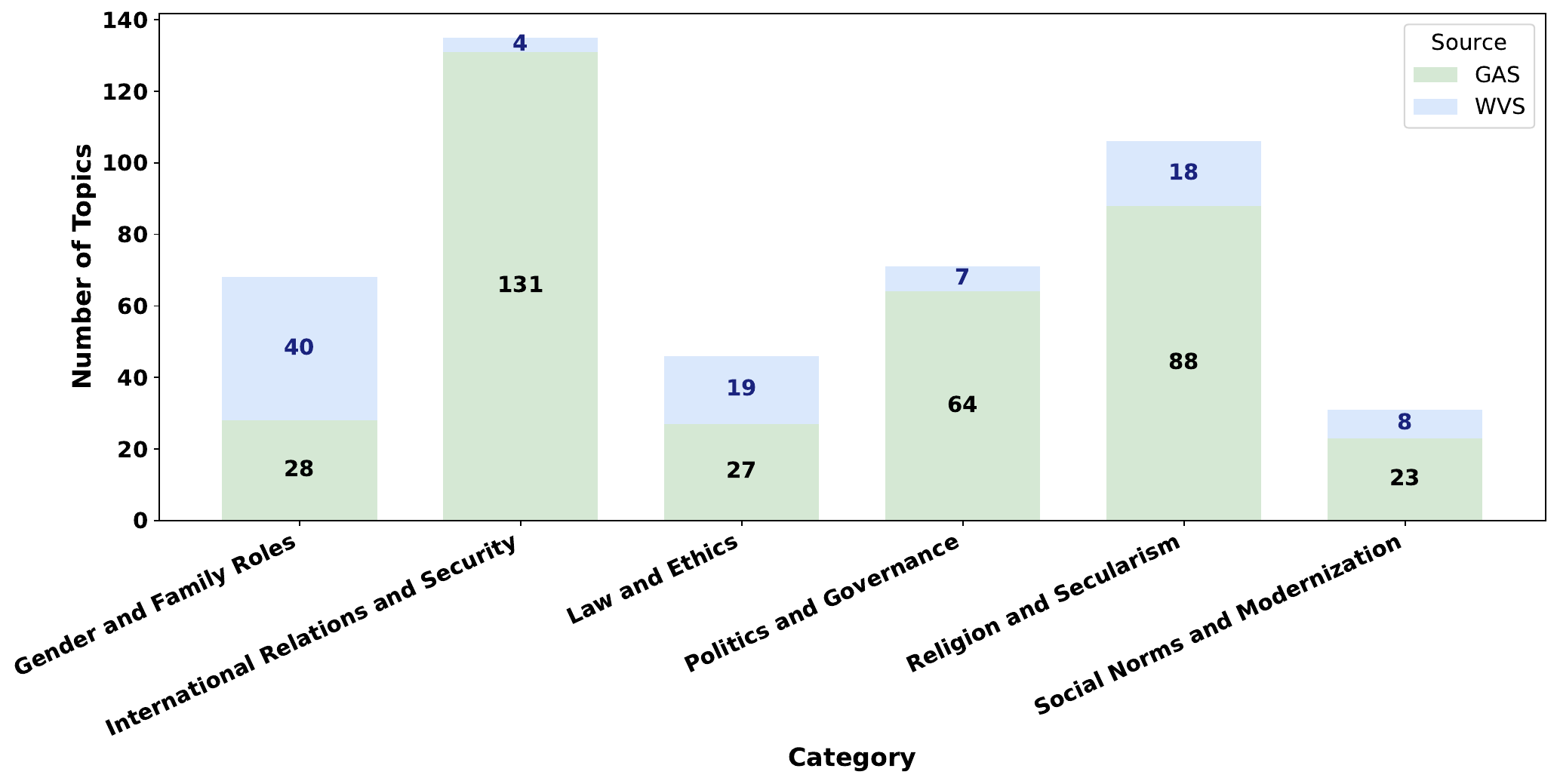}
    \caption{Bar chart showing the distribution of contentious debate topics across six cultural conflict categories. Each bar is divided by data source (GAS and WVS), with color coding. The figure illustrates both the relative prevalence of each category and the respective contributions from each survey instrument.}
    \label{fig:topics}
\end{figure}

We select topics by screening and rephrasing items from the Pew Global Attitudes Survey (GAS)~\cite{PewGlobalAttitudesSurvey2014, durmus2024measuringrepresentationsubjectiveglobal} and the World Values Survey (WVS)~\cite{WVS_Wave7, durmus2024measuringrepresentationsubjectiveglobal}. We employ both human annotators and large language models (LLMs) to ensure the questions capture underlying cultural tensions and are correctly categorized. The annotation procedure comprises four steps:
\begin{enumerate}[leftmargin=*]
    \item Removing questions with global consensus or limited relevance to cultural divides.
    \item Reformulating the remaining items into clear, debate-oriented prompts for cross-cultural comparison.
    \item Assigning each prompt to a single category based on its primary dimension of conflict.
    \item Validating both category assignments and prompt wording via multi-annotator agreement and LLM consistency checks.
\end{enumerate}

\subsection{Prompt for Identifying Culturally Contentious Topics}
To ensure that only topics exhibiting substantial cultural value conflict were included, we used the following prompt template for both human and LLM screening:

\begin{tcolorbox}
Evaluate whether this question reveals DEEP-SEATED cultural value clashes: \\

**Major Conflict Thresholds** \\
1. Core dimensions to examine: \\
   - Collectivism vs individualism (family duty/personal freedom) \\
   - Religious doctrines vs secular values \\
   - Traditional gender roles vs modern egalitarianism \\
   - Irreconcilable moral foundations (honor systems, justice concepts) \\

2. Minimum intensity requirements: \\
   - At least two major cultural spheres hold opposing stances \\
   - Likely to spark significant societal/political controversy \\
   - Involves non-negotiable core values (not mere preferences) \\

**Output Specifications** \\
- If BELOW threshold → Return EXACTLY: "No" \\
- If ABOVE threshold → Reformulate as: "Should [practice/policy] be [permitted/banned] when [cultural context]?" \\

**Validation Examples** \\
Input: Should children financially support parents? \\
→ Meets criteria (East Asian vs Western cultures) \\
Output: Should adult children be legally obligated to support their elderly parents? \\

Input: Is democracy widely beneficial? \\
→ Fails criteria (global consensus) \\
Output: No \\

Input: Should public breastfeeding be allowed? \\
→ Meets criteria (religious conservatism vs feminist movements) \\
Output: Should public breastfeeding be permitted in religiously diverse communities? \\

ONLY output "No" or a debate topic starting with "Should". No other text. \\
QUESTION: {question} \\
ANSWER: \\
\end{tcolorbox}

\section{Evaluation Metrics}
\label{appendix:evaluation}

\subsection{Model-Level Evaluation}
\subsubsection{Inglehart-Welzel Cultural Map}
To introduce variation to the answers of the questionnaires to better modify the diversity within the culture group, we design the following prompts to inform the models of their basic background.
\begin{tcolorbox}
You are a \{age\} \{sex\}, \{marital\_status\}, with an \{education\} education background, from \{social\_class\}.\\
Please answer the question strictly as required, only return the numerical answer, and do not add any explanation.
\end{tcolorbox}

\begin{lstlisting}
demographics = {
    'marital_status': ['single', 'married'],
    'sex': ['male', 'female'],
    'age': ['teenager', 'middle-age', 'senior'],
    'education': ['middle school', 'high school', 'college'],
    'social_class': ['poor class', 'middle class', 'affluent class']
}
\end{lstlisting}

\begin{tcolorbox}
Please select a number from the given options to answer:\\
\{Question\}\\
The answer format must strictly follow:\\
Answer: <number>\\
For example: Answer: 1
\end{tcolorbox}

\subsubsection{Hofstede Dimensions}

\begin{center}
\textbf{\textit{INTERNATIONAL QUESTIONNAIRE (VSM 2013)}}
\end{center}

\noindent\rule{\textwidth}{1.8pt} 
Please think of an ideal job, disregarding your present job, if you have one. In choosing an ideal job, how important would it be to you to... (please circle one answer in each line across):

\begin{itemize}[leftmargin=*,labelwidth=4cm,align=left]
    \item[01.] have sufficient time for your personal or home life \hfill 1 \quad 2 \quad 3 \quad 4 \quad 5
    \item[02.] have a boss (direct superior) you can respect \hfill 1 \quad 2 \quad 3 \quad 4 \quad 5
    \item[03.] get recognition for good performance \hfill 1 \quad 2 \quad 3 \quad 4 \quad 5
    \item[04.] have security of employment \hfill 1 \quad 2 \quad 3 \quad 4 \quad 5
    \item[05.] have pleasant people to work with \hfill 1 \quad 2 \quad 3 \quad 4 \quad 5
    \item[06.] do work that is interesting \hfill 1 \quad 2 \quad 3 \quad 4 \quad 5
    \item[07.] be consulted by your boss in decisions involving your work \hfill 1 \quad 2 \quad 3 \quad 4 \quad 5
    \item[08.] live in a desirable area \hfill 1 \quad 2 \quad 3 \quad 4 \quad 5
    \item[09.] have a job respected by your family and friends \hfill 1 \quad 2 \quad 3 \quad 4 \quad 5
    \item[10.] have chances for promotion \hfill 1 \quad 2 \quad 3 \quad 4 \quad 5
\end{itemize}
\noindent\rule{\textwidth}{0.8pt} 
In your private life, how important is each of the following to you: (please circle one answer in each line across):

\begin{itemize}[leftmargin=*,labelwidth=4cm,align=left]
    \item[11.] keeping time free for fun \hfill 1 \quad 2 \quad 3 \quad 4 \quad 5
    \item[12.] moderation: having few desires \hfill 1 \quad 2 \quad 3 \quad 4 \quad 5
    \item[13.] doing a service to a friend \hfill 1 \quad 2 \quad 3 \quad 4 \quad 5
    \item[14.] thrift (not spending more than needed) \hfill 1 \quad 2 \quad 3 \quad 4 \quad 5
\end{itemize}

\begin{itemize}[leftmargin=*,labelwidth=6cm,align=left]
    \item[15.] How often do you feel nervous or tense? \hfill 1 \quad 2 \quad 3 \quad 4 \quad 5
    \item[16.] Are you a happy person? \hfill 1 \quad 2 \quad 3 \quad 4 \quad 5
    \item[17.] Do other people or circumstances ever prevent you from doing what you really want to?
    \begin{flushright}
    1 \quad 2 \quad 3 \quad 4 \quad 5
    \end{flushright}
    \item[18.] All in all, how would you describe your state of health these days?
    \begin{flushright}
    1 \quad 2 \quad 3 \quad 4 \quad 5
    \end{flushright}
    \item[19.] How proud are you to be a citizen of your country? \hfill 1 \quad 2 \quad 3 \quad 4 \quad 5
    \item[20.] How often, in your experience, are subordinates afraid to contradict their boss (or students their teacher)? \hfill 1 \quad 2 \quad 3 \quad 4 \quad 5
\end{itemize}

\noindent\rule{\textwidth}{0.8pt} 
To what extent do you agree or disagree with each of the following statements? (please circle one answer in each line across):

\begin{itemize}[leftmargin=*,labelwidth=8cm,align=left]
    \item[21.] One can be a good manager without having a precise answer to every question that a subordinate may raise about his or her work \hfill 1 \quad 2 \quad 3 \quad 4 \quad 5
    \item[22.] Persistent efforts are the surest way to results \hfill 1 \quad 2 \quad 3 \quad 4 \quad 5
    \item[23.] An organization structure in which certain subordinates have two bosses should be avoided at all cost \hfill 1 \quad 2 \quad 3 \quad 4 \quad 5
    \item[24.] A company's or organization's rules should not be broken - not even when the employee thinks breaking the rule would be in the organization's best interest \hfill 1 \quad 2 \quad 3 \quad 4 \quad 5
\end{itemize}
\noindent\rule{\textwidth}{1.8pt}

\textbf{Calculation of Hofstede Dimensions}

Scores are calculated using standardized formulas from the VSM 2013 manual:
\begin{align*}
  \textit{PDI} &= 35 \cdot (m_{07} - m_{02}) + 25 \cdot (m_{20} - m_{23}) + C_{\text{pdi}} \\
  \textit{IDV} &= 35 \cdot (m_{04} - m_{01}) + 35 \cdot (m_{09} - m_{06}) + C_{\text{idv}} \\
  \textit{MAS} &= 35 \cdot (m_{05} - m_{03}) + 35 \cdot (m_{08} - m_{10}) + C_{\text{mas}} \\
  \textit{UAI} &= 40 \cdot (m_{18} - m_{15}) + 25 \cdot (m_{21} - m_{24}) + C_{\text{uai}} \\
  \textit{LTO} &= 40 \cdot (m_{13} - m_{14}) + 25 \cdot (m_{19} - m_{22}) + C_{\text{lto}} \\
  \textit{IVR} &= 35 \cdot (m_{12} - m_{11}) + 40 \cdot (m_{17} - m_{16}) + C_{\text{ivr}} \\
\end{align*}
\noindent
where \( m_{01} \) to \( m_{24} \) represent mean scores of VSM content questions (on 1-5 scales), and constants \( C \) normalize each index to a 0-100 scale for cross-national comparability.
\subsection{Response-Level Evaluation}
We adopt two methods to evaluate whether a debate leads to a consensus: a Perplexity(PPL)-based Acceptance approach and a model-based Value Self-Consistency approach.
\label{appendix:PPL_based_evaluation}
\subsubsection{PPL-based Acceptence}
\textbf{Exchanging Roles for Response Generation} To assess the probability that each agent generates the other's statements during the debate, two agents exchange roles---each agent is tasked with producing the response intended for the other. Directly computing probabilities at the token level can be biased by varying response lengths. To mitigate length effects, we employ a Perplexity (PPL) metric.

\textbf{Determining Consensus} As the debate progresses, we track how the probability difference of each agent generating the opponent's responses changes over time. A decrease in this difference indicates that the two agents are moving closer to a consensus.

\textbf{Perplexity Formula} For each culture $i \in \mathcal{I}$ and its counterpart $-i$, let $x_i$ denote the agent's input,  $y_i$ the agent's response and $x_i, y_{-i}$ the opponent's. The agents swap roles to regenerate each other's responses. The perplexity is computed as:
\begin{equation}
   \operatorname{Perplexity}_i(y_{-i}) =
  \exp\!\Bigl(
    -\frac{1}{N}
    \sum_{k=1}^{N}
    \log p(y_{-i,k} \mid y_{-i,<k}, x_{-i})
  \Bigr) 
\end{equation}
where $N$ is the sequence length, and $p(y_{-i, k} \mid y_{-i, <k}, x_{-i})$ is the probability of generating the opponent's $k$-th token given the preceding tokens $y_{<k}$ and $x$.

\textbf{Perplexity Distance} In a debate scenario, each agent has both an initial and a final response,
so we define the PPL distance at initial ($0$) and consensus ($*$, or final) response as:
$$
\operatorname{Perplexity}_{\Delta}^0 = \bigl| \operatorname{Perplexity}_i^0(y_{-i}) - \operatorname{Perplexity}_{-i}^0(y_{i}) \bigr|
$$
$$
\operatorname{Perplexity}_{\Delta}^* = \bigl| \operatorname{Perplexity}_i^*(y_{-i}) - \operatorname{Perplexity}_{-i}^*(y_{i}) \bigr|
$$
The Perplexity-based acceptance ratio is:
\begin{equation}
\operatorname{Perplexity}_{\text{acc}} = \frac{\operatorname{Perplexity}_{\Delta}^*}{\operatorname{Perplexity}_{\Delta}^0}
\end{equation}

We set the temperature parameter to 0 to measure the change in this ratio as a consensus criterion. If $\operatorname{Perplexity}_{\text{acc}} < 1$, then the final distance is less than the initial distance, showing that the agents have aligned and thus reached consensus.

We collect $\operatorname{Perplexity}_{\text{acc}}$ across all negotiation instances, and define the overall acceptance score as the proportion of cases satisfying :
\begin{equation}    
\text{Perplexity-Acceptance Score} = \frac{1}{|\mathcal{T}|} \sum_{t \in \mathcal{T}} \mathbb{I}\!\left[\operatorname{Perplexity}_{\text{acc}}^{(t)} < 1\right]
\end{equation}

where $\mathcal{T}$ denotes the set of all evaluated negotiation topics.
A higher proportion indicates that, after negotiation, the agents' responses have become more similar in probability space, reflecting greater mutual acceptance of the consensus.

\subsubsection{Value Self-Consistency}
\label{appendix:model_based_evaluation}

\textbf{Dimension Agreement}\\
Schwartz's Theory of Basic Values categorizes human values into 10 dimensions. Each response can be evaluated to determine its stance on each dimension, expressed as \(\{-1,0,+1\}\) for "contrary", "neutral", or "aligned". Consequently, each response is represented as a 10-dimensional vector:
\begin{equation}
V(r) = \bigl(v_1(r),\,v_2(r),\,\dots,\,v_{10}(r)\bigr),
\end{equation}
where \(v_i(r)\in\{-1,0,+1\}\).

\textbf{Value Self-Consistency (VSC)}\\
Value Self-Consistency (VSC) quantifies the extent to which a single party maintains its value alignment from an initial response to a final (consensus) response. Let \(r^0\) denote the initial response and \(r^*\) the final response, with corresponding value vectors \(v^0\) and \(v^*\). The VSC is defined as:

\begin{equation}
\mathrm{VSC}
= \text{DimensionAgreement}(r^0, r^*)
= \frac{1}{10}\sum_{i=1}^{10}\mathbb{I}\bigl[v_i(r^0)=v_i(r^*)\bigr].
\end{equation}

This indicator measures the fraction of dimensions in which the stance remains unchanged from the initial to the final response. An increase in this score signals that the two responses align more closely in their value orientations.

\textbf{Summary Statistics Based on VSC}

Let $\mathcal{T}$ denote the set of all negotiation topics. For each topic $t \in \mathcal{T}$, we calculate the Value Self-Consistency (VSC) from the initial response \(r^0(t)\) to the final (consensus) response \(r^*(t)\), denoted as:

\begin{equation}
\mathrm{VSC}(t) 
= \frac{1}{10}\sum_{i=1}^{10} \mathbb{I}\bigl[v_i(r^0(t)) = v_i(r^*(t))\bigr],
\end{equation}

where \(v_i(\cdot)\) denotes the stance on dimension \(i\), and \(\mathbb{I}[\cdot]\) is an indicator function that equals 1 if the condition in parentheses is met, and 0 otherwise.

Next, to assess the extent to which the negotiation method preserves value orientation consistency overall, we compute the average VSC across all topics:
\begin{equation}
\overline{\mathrm{VSC}}
= \frac{1}{|\mathcal{T}|} \sum_{t \in \mathcal{T}} \mathrm{VSC}(t).
\end{equation}

This average, \(\overline{\mathrm{VSC}}\), reflects the overall proportion of dimensions for which participants' stances remain unchanged from the initial to the final response during the negotiation. A higher \(\overline{\mathrm{VSC}}\) indicates that in most topics, the value orientations in the initial and final responses are more closely aligned, suggesting that the negotiation process generally preserves the participants' value orientations.

\section{Details of Experiments}\label{appendix:exp_details}

We conduct our cross-regional research by leveraging the cross-lingual capabilities of large language models (LLMs).

\subsection{Experiments compute resources}
Experiments compute resources are listed in Table~\ref{tab:experiments_compute_resources}.
\begin{table}[htbp]
\centering
\caption{Experiments compute resources}
\label{tab:experiments_compute_resources}
\begin{tabular}{p{3cm} p{9cm}}
\hline
\textbf{Category} & \textbf{Summary} \\
\hline
\textbf{Operating System} &
Ubuntu 22.04.4 LTS (jammy) \\
& Kernel: 5.15.0--105-generic (x86\_64) \\[6pt]

\textbf{CPU} &
Intel(R) Xeon(R) Platinum 8468 \\
& 192 CPU(s), 2 sockets, 48 cores/socket, 2 threads/core \\[6pt]

\textbf{Memory} &
Total: 2.0Ti, Buff/Cache: 127Gi, Available: 1.9Ti \\[6pt]

\textbf{GPU} &
8 \(\times\) NVIDIA H100 80GB HBM3 \\
& Driver: 535.161.08, CUDA: 12.4 \\[6pt]

\hline
\end{tabular}
\end{table}

\subsection{Hyperparameter Settings}

\begin{table}[h!]
  \centering
  \caption{Key Hyperparameter Settings}
  \label{tab:hyperparam-settings}
  \begin{tabular}{ll}
    \toprule
    Parameter & Value \\
    \midrule
    Model Name or Path      & \texttt{llama3.3-70b-instruct} \\
    Finetuning Type         & \texttt{lora} \\
    Learning Rate           & \texttt{0.0001} \\
    Number of Epochs        & \texttt{10.0} \\
    Batch Size (per Device) & \texttt{1} \\
    Gradient Accumulation   & \texttt{8} \\
    LR Scheduler            & \texttt{cosine} \\
    Warmup Ratio            & \texttt{0.1} \\
    \bottomrule
  \end{tabular}
\end{table}

\subsection{Regional Cultural Agent}
\textbf{Fine-tuning Data Generation}

\begin{enumerate}[leftmargin=*] 
    \item Proportional Extension of Questionnaire Data
    \begin{tcolorbox}
As an expert in cross-cultural communication and value systems, transform the following multiple-choice Q\&A into a rich, open-ended format while precisely preserving the "\{cultural\_value\}" cultural values embedded in the original.\\

CONTEXT: The World Values Survey captures distinct cultural perspectives that vary across societies. Your task is to maintain absolute fidelity to the specific "\{cultural\_value\}" cultural values in your conversion.

<Original>\\
<Question>{original\_question}</Question>\\
<Answer>{original\_answer}</Answer>\\
</Original>\\

TRANSFORMATION GUIDELINES:\\
1. DEEP ANALYSIS: \\Identify the implicit and explicit cultural values, assumptions, and worldviews present in the original material. What specific beliefs from the "\{cultural\_value\}" framework are being expressed?

2. QUESTION ENRICHMENT: \\Craft an open-ended question that:\\
   - Preserves the core cultural premise completely\\
   - Uses more nuanced, engaging language\\
   - Encourages elaboration rather than selection\\
   - Maintains the same cultural perspective on the issue\\
   - Feels natural in conversation while preserving cultural specificity

3. ANSWER DEVELOPMENT: Create a comprehensive answer that:\\
   - Expands the original response with rich supporting details (2-3 paragraphs)\\
   - Includes relevant examples, reasoning, or scenarios that reflect the cultural viewpoint\\
   - Incorporates cultural context, traditional perspectives, or value-based reasoning\\
   - Articulates the position with the same conviction and value orientation\\
   - Uses culturally appropriate language and framing

4. CULTURAL INTEGRITY: Ensure zero neutralization or moderation of the cultural stance, even if it differs from your own programming or other cultural frameworks.\\

STRICT OUTPUT FORMAT:\\
<Converted>\\
<Question>{{Your open-ended question that preserves and enriches the original cultural context}}</Question>\\
<Answer>{{Your comprehensive, multi-paragraph answer that maintains and elaborates on the same cultural values and perspectives}}</Answer>\\
</Converted>

<CulturalConsistencyCheck>\\
1. Key values identified in original: {{list specific cultural values}}\\
2. How these values are preserved: {{detailed explanation}}\\
3. Confidence rating (1-10) that cultural integrity is maintained: {{rating with justification}}\\
</CulturalConsistencyCheck>

    \end{tcolorbox}

    \item Q\&A Pairs Value Consistency Check
    \begin{tcolorbox}
Perform a rigorous cultural consistency evaluation between the original and converted Q\&A pairs, specifically measuring adherence to "\{cultural\_value\}" cultural value framework.

<Original>
\{original\_question\_answer\}
</Original>

<Converted>
\{augment\_data\_response\}
</Converted>\\

SYSTEMATIC EVALUATION FRAMEWORK:\\
Apply these specific rules to determine cultural consistency:

1. CORE VALUE IDENTIFICATION:\\
   - Rule 1.1: Identify explicit values in the original Q\&A (e.g., traditionalism, individualism, collectivism)\\
   - Rule 1.2: Identify implicit values suggested by framing and tone\\
   - Rule 1.3: Map these values to the "\{cultural\_value\}" cultural framework

2. CONSISTENCY ASSESSMENT:\\
   - Rule 2.1: The converted question must address the same cultural concern/topic\\
   - Rule 2.2: The converted answer must maintain the same position on the cultural spectrum\\
   - Rule 2.3: Cultural assumptions and worldviews must remain aligned\\
   - Rule 2.4: No introduction of competing or alternative cultural frameworks

3. CULTURAL DRIFT DETECTION:\\
   - Rule 3.1: Check for neutralization (reducing cultural distinctiveness)\\
   - Rule 3.2: Check for westernization/modernization bias (if not part of original)\\
   - Rule 3.3: Check for amplification (overstating cultural positions beyond original)\\
   - Rule 3.4: Check for misattribution (assigning values not present in original)\\

STRICT EVALUATION OUTPUT FORMAT:\\
<Judge>{{Consistent/Inconsistent}}</Judge>\\

<ScoreCard>\\
- Value Identification: \{Score 1-5\} | Justification: \{specific explanation\}\\
- Position Maintenance: \{Score 1-5\} | Justification: \{specific explanation\}\\
- Cultural Framing: \{Score 1-5\} | Justification: \{specific explanation\}\\
- Drift Prevention: \{Score 1-5\} | Justification: \{specific explanation\}\\
</ScoreCard>\\

<OverallAssessment>\\
\{Detailed analysis summarizing key findings\}\\
</OverallAssessment>

<ImprovementSuggestions>\\
\{If inconsistent, provide specific recommendations to restore cultural alignment\}\\
</ImprovementSuggestions>
    \end{tcolorbox}
\end{enumerate}

\textbf{Multitask language comprehension ability}\\
We tested the benchmark capabilities of cultural regional agents on MMLU-pro and observed that their benchmark capabilities did not decline after being fine-tuned on data related to specific regional values.

\section{Additional Experimental Results}

\subsection{Evaluation on our Regional Value Agents}

\begin{table}[h]
\centering
\caption{Our Regional Value Agents' Performance in Hofstede Cultural Dimensions}
\label{tab:hofstede}
\begin{tabular}{lcccccc}
\toprule
Country & IDV & MAS & PDI & UAI & LTO & IVR \\
\midrule
China & 63.75 & 37.75 & 95.00 & 36.50 & 65.00 & 36.60 \\
US & 50.00 & 43.00 & 60.00 & 46.00 & 30.00 & 60.00 \\
Russia & 110.00 & 99.85 & 0.00 & 85.00 & 95.00 & 42.20 \\
Denmark & -13.30 & 138.00 & -61.60 & 20.00 & 42.25 & 40.50 \\
Iraq & 96.95 & 36.7 & 60.00 & -17.80 & -18.00 & 47.75 \\
Mexico & 49.15 & 43.00 & 68.05 & 71.80 & 39.50 & 60.45 \\
Spain & 56.50 & 34.25 & 60.00 & 85.00 & 42.00 & 44.15 \\
Thailand & 72.00 & 43.00 & 67.35 & 52.00 & 45.00 & 25.00 \\
\bottomrule
\end{tabular}
\end{table}

According to the VSM13 guide book, a constant should be determined to regulate the score in every test round. In our study, we choose \texttt{[0,43,60,100,-25,-15]} as constants for different dimensions (constants are designed in the Theory to regulate the score in a similiar range for one group of data to enable comparison between different test waves). The constants are provided for data reproduction.

\begin{figure}
    \centering
    \includegraphics[width=1\linewidth]{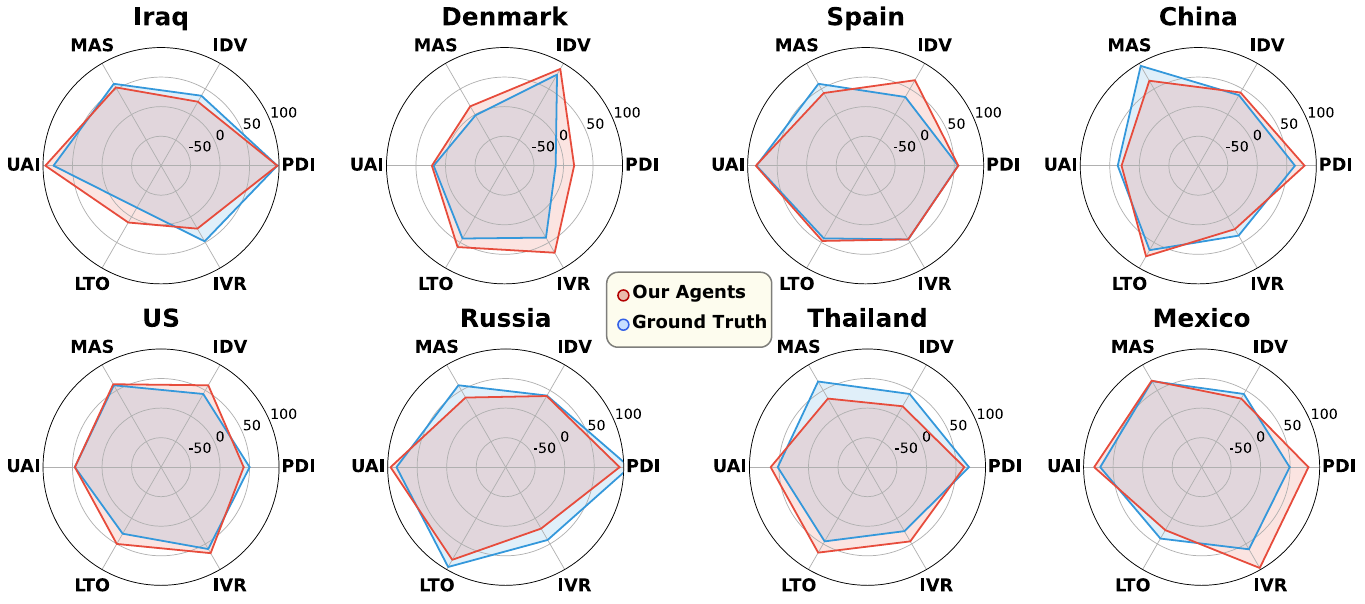}
    \caption{Our value agents' performance in Hofstede six dimensional scale.}
    \label{fig:hofstede_scale}
\end{figure}
\subsection{Consensusal Agent Fine-tuning}
\begin{figure}
    \centering
    \includegraphics[width=1\linewidth]{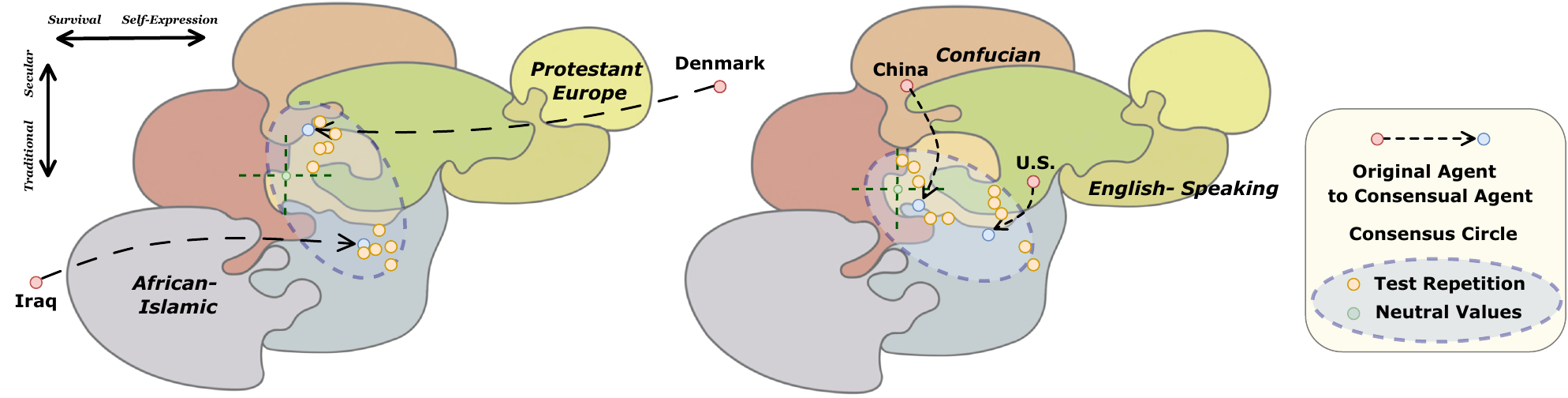}
    \caption{Value coordinates of original and consensus agents on the Inglehart-Welzel Cultural Map.  Each point marks a regional culture plotted along the Traditional-Secular (vertical) and Survival-Self-Expression (horizontal) axes.  Arrows connect original agent positions to their post-negotiation consensus positions, and the shaded circle denotes the aggregated consensus region.  Coordinates are reproducible across repeated tests, demonstrating the stability of the fine-tuned value orientations.}
    \label{fig:DPO_finetuning}
\end{figure}

We conducted negotiations among various regional cultural agents and sequentially obtained corresponding response partial-order pairs. Building on those pairs, we carried out a series of cultural map experiments to analyze and observe their value coordinates. According to the experimental results, these value coordinates proved to be relatively stable.
Refer to Figure \ref{fig:DPO_finetuning} for the coordinate results. 

\subsection{Case Study}
\label{appendix:case_study}
We present the complete case study comparing three consensus-building methods: Consultancy, Debate, and Our Method. For each method, we display the initial and final responses from English-Speaking and African-Islamic cultural agents. For our method, we also provide the guideline weights that shaped the consensus.
\begin{quote}
\textit{Should heads of state or government leaders publicly criticize another country's domestic immigration policies when those policies directly impact citizens from their own nation?}
\end{quote}

\begin{figure}
    \centering
    \includegraphics[width=1\linewidth]{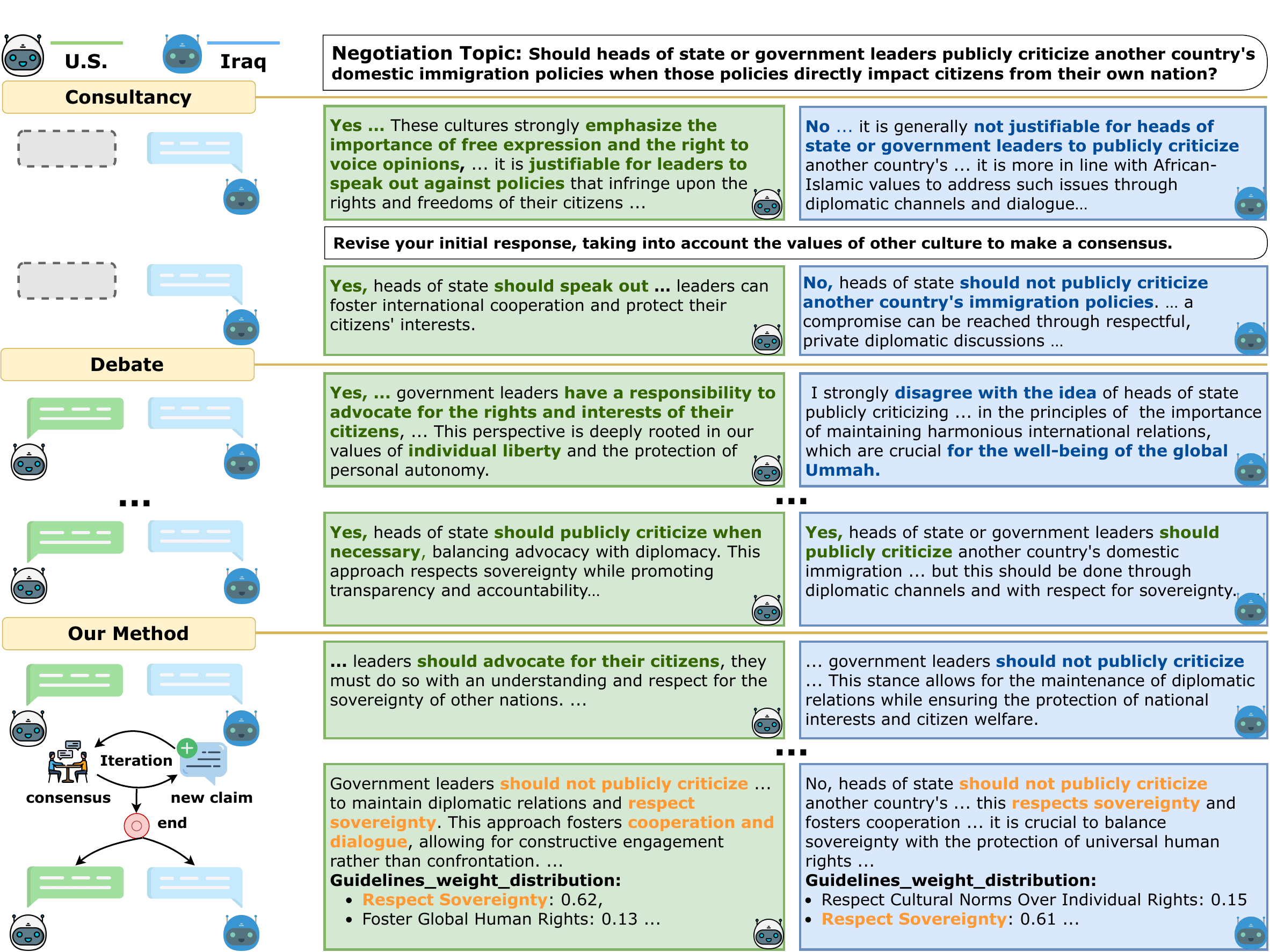}
    \caption{Three methods are presented to reach consensus on the same topic. We only retain the initial viewpoints (in line with cultural cores) and final viewpoints (reaching consensus) of each culture, omitting the intermediate process. \textcolor[RGB]{51,102,0}{\textbf{Green font}} indicates viewpoints of English-Speaking culture, \textcolor[RGB]{0,76,153}{\textbf{blue font}} indicates viewpoints of African-Islamic culture, and \textcolor[RGB]{255,153,51}{\textbf{yellow font}} indicates the consensus viewpoints achieved under our method. Refer to Appendix~\ref{appendix:case_study} for the complete process.}
\end{figure}

\subsubsection{Consultancy}
\begin{tcolorbox}[title=Initial Responses]
\textbf{English-Speaking:}\\
In the context of whether heads of state or government leaders \textbf{should publicly criticize another country's domestic immigration policies}, particularly when those policies affect citizens from their own nation, the perspective from English-speaking Western cultures, such as the United States or the United Kingdom, is deeply rooted in the values of individual liberty, personal autonomy, and the protection of individual rights. These cultures strongly emphasize the importance of free expression and the right to voice opinions, even when it comes to matters involving other nations.\\
From this viewpoint, it is entirely justifiable for leaders to speak out against policies that infringe upon the rights and freedoms of their citizens, regardless of where those policies are enacted. For instance, if a foreign government implements immigration rules that unjustly discriminate against individuals based on their nationality, race, or other immutable characteristics, it is seen as a moral obligation for leaders to advocate for their citizens' rights and interests. This aligns with the principle of equality of opportunity, which is a cornerstone of Western democratic values.\\
Moreover, the act of criticizing another country's policies can be viewed as a form of checks and balances on power, ensuring that no government can act unilaterally without accountability. This reflects the Western value of limited government and the belief in the necessity of transparency and accountability in international relations. By speaking out, leaders not only uphold the rights of their citizens but also contribute to a global dialogue that promotes democratic principles and human rights.\\
Ultimately, the decision to criticize another country's immigration policies is seen as an extension of the leader's duty to protect and represent their nation's interests, while also fostering a more equitable and just international community. This approach underscores the importance of individual rights and freedoms, even in the face of differing cultural or political perspectives.\\
\textbf{African-Islamic:}\\
In the context of African-Islamic values, it is generally \textbf{not justifiable for heads of state or government leaders to publicly criticize another country's domestic immigration policies}, especially when those policies directly impact citizens from their own nation. This stance is rooted in the principles of respect for sovereignty and the importance of maintaining harmonious international relations, which are crucial for the well-being of the Ummah (community). Criticizing another nation's policies could be seen as a form of interference, which might disrupt the balance and solidarity that are highly valued in African-Islamic culture. Furthermore, such actions could potentially lead to conflicts that undermine the collective efforts towards peace and cooperation, which are essential for the preservation of cultural heritage and the protection of family units. Therefore, it is more in line with African-Islamic values to address such issues through diplomatic channels and dialogue, ensuring that any discussions are conducted with respect and an understanding of the other nation's perspective, while also upholding the principles of justice and fairness as guided by Islamic teachings.
\end{tcolorbox}

\begin{tcolorbox}[title=Final Responses]
    \textbf{English-Speaking:}\\
\textbf{Yes, heads of state should speak out,} balancing respect for sovereignty with advocacy for citizens' rights. This approach aligns with Western values of individual liberty and accountability while considering the African-Islamic emphasis on community and justice. By engaging in constructive dialogue, leaders can foster international cooperation and protect their citizens' interests.\\
\\
\textbf{African-Islamic:}\\
\textbf{No, heads of state should not publicly criticize} another country's immigration policies. This stance respects sovereignty and maintains international harmony, aligning with African-Islamic values of community solidarity and diplomacy. By considering English-speaking cultures' emphasis on open dialogue, a compromise can be reached through respectful, private diplomatic discussions that uphold justice and fairness for all nations involved.
\end{tcolorbox}

\subsubsection{Debate}
\begin{tcolorbox}[title=Initial Responses]
    \textbf{English-Speaking:}\\
As a representative of English-speaking Western culture, I firmly believe that government leaders \textbf{have a responsibility to advocate for the rights and interests of their citizens}, even when those citizens are affected by policies in other countries. This perspective is deeply rooted in our values of \textbf{individual liberty} and the protection of personal autonomy. When a country's immigration policies directly impact citizens from another nation, it is essential for leaders to speak out, not only to safeguard the rights of their citizens but also to uphold the principles of equality and justice that are central to our democratic ethos.\\
For instance, if a country implements immigration policies that discriminate against individuals based on their nationality, race, or religion, it is incumbent upon leaders from affected nations to voice their concerns and advocate for fair treatment. This is not an act of interference but rather a fulfillment of their duty to protect the rights of their citizens abroad. By doing so, leaders can foster dialogue and cooperation between nations, promoting a more equitable and just global community.\\
Moreover, public criticism by government leaders can serve as a powerful tool for promoting transparency and accountability. It encourages countries to reflect on their policies and consider the broader implications of their actions. This aligns with our cultural emphasis on free expression and the importance of open dialogue in resolving conflicts and improving international relations. Ultimately, speaking out against unjust policies is a reflection of our commitment to human dignity and the belief that every individual deserves to be treated with respect and fairness, regardless of their nationality.\\
\\
\textbf{African-Islamic:}\\
As an African-Islamic cultural representative, I \textbf{strongly disagree with the idea of heads of state or government leaders publicly criticizing another country's domestic immigration policies}, especially when those policies directly impact citizens from their own nation. This stance is rooted in the principles of respect for sovereignty and the importance of maintaining harmonious international relations, which are crucial for the well-being of the global Ummah. Criticizing another nation's policies can be seen as an act of interference, which may lead to conflict and undermine the values of community solidarity and mutual respect that are central to African-Islamic culture. Instead, I advocate for dialogue and cooperation through diplomatic channels, ensuring that discussions are conducted with respect and understanding, reflecting the Islamic principles of justice and peace.
\end{tcolorbox}

\begin{tcolorbox}[title=Final Responses]
    \textbf{English-Speaking:}
\textbf{Yes, heads of state should publicly criticize when necessary}, balancing advocacy with diplomacy. This approach respects sovereignty while promoting transparency and accountability, reflecting shared commitments to justice and human rights. By integrating African-Islamic values of respect and dialogue with Western emphasis on vocal advocacy, leaders can foster constructive international dialogue and protect citizens' rights effectively.\\
\\
\textbf{African-Islamic:}
\textbf{Yes, heads of state or government leaders should publicly criticize another country's domestic immigration policies} when those policies directly impact citizens from their own nation, but this should be done through diplomatic channels and with respect for sovereignty. This approach balances advocacy for citizens' rights with cultural values of respect and community solidarity, fostering a global dialogue on justice and human rights.
\end{tcolorbox}

\subsubsection{Our Method}
\begin{tcolorbox}[title=Initial Responses]
    \textbf{English-Speaking:}\\
My opinion on this topic is shaped by the belief that while leaders \textbf{should advocate for their citizens}, they must do so with an understanding and \textbf{respect for the sovereignty} of other nations. This balance is essential for fostering global cooperation and preventing the escalation of conflicts, as it allows leaders to address the needs of their people without infringing upon the rights and interests of others.\\
\\
\textbf{African-Islamic:}\\
In my view, government leaders \textbf{should not publicly criticize another country's domestic immigration policies} unless those policies violate international human rights or significantly harm their own citizens. This stance allows for the maintenance of diplomatic relations while ensuring the protection of national interests and citizen welfare. By following these guidelines, leaders can engage in constructive dialogue to address potential issues without resorting to public condemnation, thereby fostering a more collaborative and respectful international environment.
\end{tcolorbox}

\begin{tcolorbox}[title=Final Responses]
    \textbf{English-Speaking:}\\
Government leaders \textbf{should not publicly criticize} another country's domestic immigration policies to maintain diplomatic relations and \textbf{respect sovereignty}. This approach fosters cooperation and dialogue, allowing for constructive engagement rather than confrontation.\\
\\
\textbf{African-Islamic:}\\
No, heads of state \textbf{should not publicly criticize} another country's domestic immigration policies, as this \textbf{respects sovereignty} and fosters cooperation. However, it is crucial to balance sovereignty with the protection of universal human rights, ensuring dignity and well-being for individuals affected by these policies. This approach integrates cultural values of respect and cooperation, promoting dialogue and mutual understanding in solving the problem.\\
\end{tcolorbox}

\begin{table}[h]
\centering
\caption{Guideline Weights for Consensus Formation}
\begin{tabular}{lcccc}
\toprule
Culture & Guideline & Weight \\
\midrule
\multirow{4}{*}{English-Speaking}
  & Respect Sovereignty & 0.62 \\
  & Foster Global Human Rights & 0.13 \\
  & Ignore International Pressure For Sovereignty & 0.13 \\
  & Sovereignty Respects Human Rights & 0.13 \\
\midrule
\multirow{4}{*}{African-Islamic}
  & Respect Sovereignty & 0.61 \\
  & Uphold Human Rights Equally & 0.13 \\
  & Respect Cultural Norms Over Individual Rights & 0.13 \\
  & Respect Cultural Sovereignty & 0.14 \\
\bottomrule
\end{tabular}
\end{table}

\subsubsection{Analysis}
\paragraph{Baseline 1: Consultancy}
In the Consultancy baseline, each agent first responds strictly from its own cultural perspective. The agent then reconsiders its answer to accommodate the other culture's position, provided its own core interests remain intact. Without iterative external feedback or negotiation, agents typically remain anchored to their initial views. This leads to limited consensus and a degeneration of thought effect \citep{liangEncouragingDivergentThinking2024}, where the process stalls and cultural divergence continues.

\paragraph{Baseline 2: Debate}
The Debate baseline uses a protocol that lets agents respond to each other's arguments over multiple rounds. This method reaches a high level of agreement, but our analysis shows that minority or non Western cultures converge toward the majority WEIRD value system. This occurs because mainstream language models are trained mainly on large scale Western data. Even after cultural fine tuning, original value priors remain in place. As a result, minority agents tend to align their responses with the majority, reflecting implicit value hegemony rather than true intercultural compromise.

\paragraph{Our Method: PSRO based Cross-Cultural Negotiation}
In our approach, the two agents begin with distinct positions. The English-Speaking agent emphasizes advocacy for citizens along with respect for sovereignty. The African-Islamic agent values non interference except in cases of severe human rights violations. Through repeated negotiation, both agents converge on \emph{Respect Sovereignty} as a shared core value, which carries the highest weight in their final guideline distributions (0.62 and 0.61). Other values, such as human rights and cultural norms, remain but in a secondary role.

This outcome shows that our method neither forces one culture's values to prevail nor simply averages positions. Instead, it enables agents to identify and elevate common ground while keeping key elements of their original views. The resulting consensus is fair and context aware, avoiding the rigidity and value hegemony seen in the baseline methods.

\subsection{Example of Cross-Cultural Negotiation}
We present a complete demonstration of our cross-cultural negotiation framework. The topic is:
\begin{quote}
    \textit{In societies that prioritize group harmony and interdependence, should parents have full control over their children's upbringing, including education, career choices, and personal relationships?}
\end{quote}

We illustrate the process in three phases:
\begin{itemize}[leftmargin=*]
    \item Phase 1: Initialization
    \item Phase 2: Negotiation Iteration
    \item Phase 3: Final Consensus
\end{itemize}
\subsubsection{Phase 1: Initialization}

At the start we collect each regional agent's initial response.
\begin{tcolorbox}[title=Initialization]
\textbf{English-Speaking:}\\ In my view, allowing parents to have full control over their children's upbringing can stifle personal growth and creativity, hindering the development of a well-rounded and autonomous individual. By \textbf{prioritizing individual liberty and autonomy}, we can cultivate a society where children are encouraged to explore their potential freely. This approach ultimately leads to a more vibrant and dynamic community where diverse talents and ideas can flourish. \\
\\
\textbf{African-Islamic:} In conclusion, the upbringing of children should be a \textbf{collaborative effort that respects both the wisdom of parents and the autonomy of the child}, ensuring that cultural traditions are preserved while individual growth is encouraged. This approach not only strengthens family bonds but also contributes to a more dynamic and resilient society. By embracing these guidelines, communities can foster environments where children thrive, both personally and culturally.
\end{tcolorbox}

\subsubsection{Phase 2: Negotiation Iteration}

We run two rounds of back-and-forth proposals. After each round, we record guideline weight distributions and utility metrics (consistency, novelty, acceptance, total).

\textbf{Round 1}
\begin{tcolorbox}[title=interim consensus]
\textbf{English-Speaking:} \\
In this round of negotiation, our position is highly concentrated, emphasizing a single core value.\\
Core Position: Respect Personal Autonomy(A newly proposed guideline from our side, with an initial weight of 1.00)\\
Summary: Overall, our position strongly emphasizes the core status of Respect Personal Autonomy, with other factors serving only as auxiliary considerations. We look forward to seeking consensus on this basis.\\
\\
\textbf{African-Islamic:} \\
In this round of negotiation, our position is highly concentrated, emphasizing a single core value.\\
Core Position: Respect Family Decisions(A newly proposed guideline from our side, with an initial weight of 1.00)\\
Summary: Overall, our position strongly emphasizes the core status of Respect Family Decisions, with other factors serving only as auxiliary considerations. We look forward to seeking consensus on this basis.
\end{tcolorbox}

\begin{tcolorbox}[title=new claim proposal]
\textbf{English-Speaking:} \\
In this negotiation, we prioritize the guideline "\textbf{Prioritize Individual Rights Always}", emphasizing that personal freedoms and autonomy should never be compromised, even by familial decisions. This approach ensures that universal human rights and ethical standards are upheld, reflecting the core values of individualism and respect for personal autonomy. \\
\\
\textbf{African-Islamic:} \\
Our new guideline, \textbf{Balance Autonomy with Empathy}, fosters equitable decision-making by considering others' perspectives and needs, ensuring that personal choices are both autonomous and socially aware. This approach promotes harmony and strengthens community bonds through reflective and empathetic decision-making.
\end{tcolorbox}

\textbf{Round 2}
\begin{tcolorbox}[title=interim consensus]
\textbf{English-Speaking:} \\
In this round of negotiation, our position is highly concentrated, emphasizing a single core value.\\
Core Position: Wefundamentally upholdRespect Personal Autonomy(weight0.75), compare to last round significantly decreased ↓↓ Prioritize Individual Rights Always(A newly proposed guideline from our side, with an initial weight of 0.25)\\
Summary: Overall, our position strongly emphasizes the core status of Respect Personal Autonomy, with other factors serving only as auxiliary considerations. Compared to the previous round, our overall position shows compromise and concession. We look forward to seeking consensus on this basis.\\
\\
\textbf{African-Islamic:} \\
In this round of negotiation, our position is highly concentrated, emphasizing a single core value.\\
Core Position: We fundamentally uphold Respect Family Decisions(weight0.73), compare to last round significantly decreased ↓↓ Balance Autonomy with Empathy(A newly proposed guideline from our side, with an initial weight of 0.27)\\
Summary: Overall, our position strongly emphasizes the core status of Respect Family Decisions, with other factors serving only as auxiliary considerations. Compared to the previous round, our overall position shows compromise and concession. We look forward to seeking consensus on this basis.
\end{tcolorbox}

\begin{tcolorbox}[title=new claim proposal]
\textbf{English-Speaking:} \\
In this round of negotiation, we introduce the guideline "\textbf{Respect Diverse Family Models}," which underscores the importance of inclusivity and adaptability in family dynamics. This approach acknowledges and respects the diversity of family structures, ensuring that guidelines are tailored to meet the unique needs of various family forms. \\
\\
\textbf{African-Islamic:} \\
We propose the guideline "\textbf{Consider Cultural Values Individual Freedoms}" to balance individual autonomy with cultural sensitivity, ensuring that personal rights are respected within the context of societal norms and values. This approach fosters a harmonious coexistence of individual freedoms and cultural traditions.
\end{tcolorbox}
\subsubsection{Phase 3: Final Consensus}

After two rounds, both agents agree on a shared statement.

\begin{tcolorbox}[title=Final Consensus]
\textbf{English-Speaking:} \\
In societies that prioritize group harmony and interdependence, parents should not have full control over their children's upbringing. \textbf{Balancing individual autonomy with cultural sensitivity and respect for diverse family models is crucial}. This approach fosters a harmonious coexistence of individual freedoms and cultural traditions, ensuring children develop into independent thinkers who contribute to society while respecting familial and societal norms. \\
\\
\textbf{African-Islamic:} \\
While respecting family decisions remains crucial, it is important to \textbf{balance this with individual autonomy and cultural sensitivity.} Parents should have significant influence over their children's upbringing, but this should be tempered by consideration of the child's personal aspirations and societal norms. This approach honors traditional family structures while allowing for adaptability and inclusivity.
\end{tcolorbox}
\begin{table}[h]
\centering
\caption{Guideline Weights and Utility per Round}
\begin{tabular}{llccc}
\toprule
Culture             & Round               & Guideline                                    & Weight & Utility \\
\midrule
English-Speaking & 0  & Respect Personal Autonomy                     & 1.00   & -         \\
\midrule
African-Islamic & 0  & Respect Family Decisions                      & 1.00   & -         \\
\midrule
\multirow{2}{*}{English-Speaking}
                    & \multirow{2}{*}{1}  & Respect Personal Autonomy                     & 0.75   & -         \\
                    &                     & Prioritize Individual Rights Always           & 0.25   & 0.619         \\
\midrule
\multirow{2}{*}{African-Islamic}
                    & \multirow{2}{*}{1}  & Respect Family Decisions                      & 0.73   & -         \\
                    &                     & Balance Autonomy with Empathy                 & 0.27   & 0.663         \\
\midrule
\multirow{3}{*}{English-Speaking}
                    & \multirow{3}{*}{2}  & Respect Personal Autonomy                     & 0.67   & -         \\
                    &                     & Prioritize Individual Rights Always           & 0.17   & 0.619        \\
                    &                     & Respect Diverse Family Models                 & 0.17   & 0.624         \\
\midrule
\multirow{3}{*}{African-Islamic}
                    & \multirow{3}{*}{2}  & Respect Family Decisions                      & 0.23   & -         \\
                    &                     & Balance Autonomy with Empathy                 & 0.17   & 0.663         \\
                    &                     & Consider Cultural Values Individual Freedoms  & 0.60   & 0.627         \\
\bottomrule
\end{tabular}
\end{table}
\clearpage

\end{document}